\RequirePackage{tikz}
\PassOptionsToPackage{hidelinks}{hyperref}

\documentclass[sn-basic,iicol]{sn-jnl}

\usepackage[english]{babel}
\usepackage[utf8]{inputenc}
\usepackage[T1]{fontenc}
\usepackage{amsmath}
\usepackage{amsfonts}
\usepackage{amssymb}
\usepackage{graphicx}
\usepackage{booktabs}
\usepackage{url}
\usepackage{array}
\usepackage{multirow}
\usepackage{tabularx}
\usepackage{subfig}
\usepackage{relsize}
\usepackage{makecell}
\usepackage{color}
\usepackage{colortbl}
\usepackage{grffile}

\usepackage{tikz}
\usetikzlibrary{shapes.geometric, arrows}
\usepackage{pgfplots}
\pgfplotsset{compat=1.9}

\usepackage{hyperref}
  \hypersetup{
    citecolor=black,
    linkcolor=black,   
    urlcolor=black
  }

\graphicspath{{./figures/}}

\newcommand{\cmmnt}[1]{}

\definecolor{drawioblue}{HTML}{DAE8FC}

\tikzstyle{startstop} = [
rectangle, 
rounded corners, 
minimum width=3cm, 
minimum height=1cm, 
text centered, 
draw=black, 
fill=drawioblue 
]
\tikzstyle{process} = [rectangle, minimum width=3cm, minimum height=1cm, text centered, draw=black]
\tikzstyle{data} = [parallelogram, minimum width=3cm, minimum height=1cm, text centered, draw=black]
\tikzstyle{arrow} = [thick,->,>=stealth]

\newcommand{\footurl}[1]{\footnote{\url{#1}}}

\newcommand\raisepunct[1]{\,\mathpunct{\raisebox{0.5ex}{#1}}}

\jyear{2023}%

\theoremstyle{thmstyleone}%
\theoremstyle{thmstyletwo}%

\theoremstyle{thmstylethree}%

\definecolor{Gray}{gray}{0.9}

\definecolor{Gray}{gray}{0.9}

\begin{document}

\tolerance=999
\sloppy


\title[ ]{Identification of Deforestation Areas in the Amazon Rainforest Using Change Detection Models}

\author*[1]{\fnm{Christian Massao} \sur{Konishi}}\email{christian.konishi@gmail.com}

\author*[1]{\fnm{Helio} \sur{Pedrini}}\email{helio@ic.unicamp.br}

\affil*[1]{\orgdiv{Institute of Computing}, \orgname{University of Campinas}, \orgaddress{\street{Av. Albert Einstein, 1251}, \city{Campinas-SP}, \postcode{13083-852}, \state{S\~ao Paulo}, \country{Brazil}}}

\abstract{The preservation of the Amazon Rainforest is one of the global priorities in combating climate change, protecting biodiversity, and safeguarding indigenous cultures. The Satellite-based Monitoring Project of Deforestation in the Brazilian Legal Amazon (PRODES), a project of the National Institute for Space Research (INPE), stands out as a fundamental initiative in this effort, annually monitoring deforested areas not only in the Amazon but also in other Brazilian biomes. Recently, machine learning models have been developed using PRODES data to support this effort through the comparative analysis of multitemporal satellite images, treating deforestation detection as a change detection problem. However, existing approaches present significant limitations: models evaluated in the literature still show unsatisfactory effectiveness, many do not incorporate modern architectures, such as those based on self-attention mechanisms, and there is a lack of methodological standardization that allows direct comparisons between different studies. In this work, we address these gaps by evaluating various change detection models in a unified dataset, including fully convolutional models and networks incorporating self-attention mechanisms based on Transformers. We investigate the impact of different pre- and post-processing techniques, such as filtering deforested areas predicted by the models based on the size of connected components, texture replacement, and image enhancements; we demonstrate that such approaches can significantly improve individual model effectiveness. Additionally, we test different strategies for combining the evaluated models to achieve results superior to those obtained individually, reaching an F1-score of 80.41\%, a value comparable to other recent works in the literature.}


\maketitle

\section{Introduction}  

\begin{figure}[!htb]
\centering
\includegraphics[width=\linewidth]{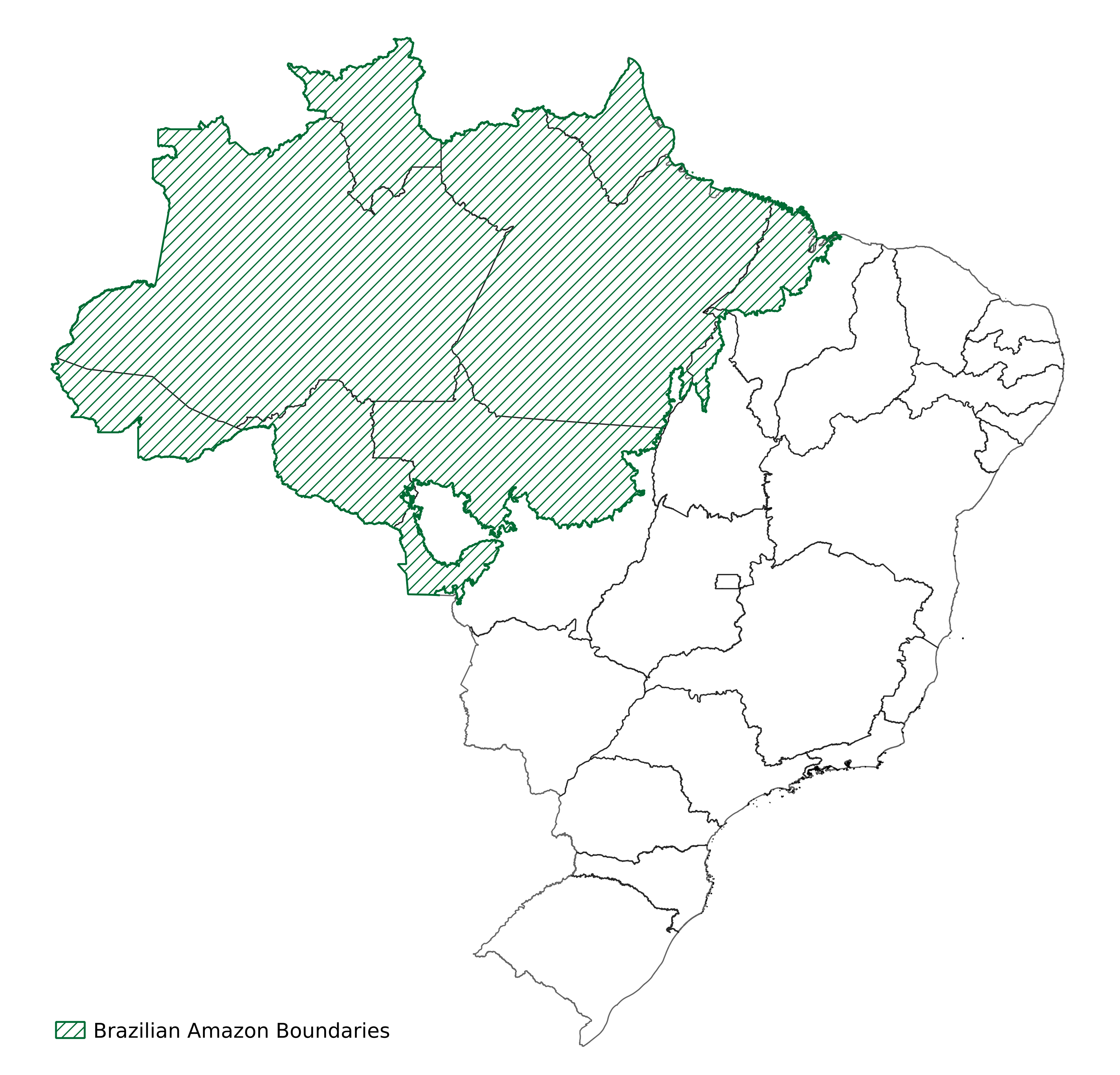}
\caption{Map of the coverage area of the Brazilian Amazon biome. Data obtained from~\citep{prodes}.}
\label{fig:biome}
\end{figure}

The deforestation of the Amazon Rainforest is a central issue in climate change discussions. This biome spans approximately 7 million square kilometers (Figure~\ref{fig:biome}) and plays a critical role in global ecological balance. While its preservation could be justified solely by its immense biodiversity, the rainforest also sustains entire communities as a source of income and provides numerous components with commercial potential.

Furthermore, the forest is home to and an essential part of the culture of approximately 180 indigenous groups in the Brazilian region alone~\citep{amazon_disease,amazon_population}. Despite this, the estimated deforestation in 2020 reached 11,088 km$^2$, the highest rate in the past decade~\citep{2020deforestation}.

A key part of biome preservation efforts lies in monitoring. For this purpose, Brazil's National Institute for Space Research (INPE) operates the DETER and PRODES projects (Project for Monitoring Deforestation in the Legal Amazon by Satellite), which identify deforested areas through satellite imagery~\citep{prodes,deforestation_success}. These data enable estimation of deforested areas, alerts about ongoing destruction, and the development of public policies to combat deforestation. 

The PRODES project specifically reports regions affected by deforestation from one year to the next, allowing mapping of each deforested area with its occurrence year. The project primarily uses imagery from Landsat Mission satellites~\citep{landsat}, made available by the United States Geological Survey (USGS).

The effort required by the PRODES project is costly, time-consuming, and relies on manual work by multiple specialists~\citep{maretto21}. An alternative to automate this process is to treat it as a change detection problem, comparing two images from different years and mapping the pixel-level alterations. 

These models aim to produce a change map (Figure~\ref{fig-method:train_samples}), also referred to as a mask, where each activated pixel corresponds to a temporal change and, furthermore, represents a new deforested area~\citep{cd_review_2022}.

This work aims to implement and evaluate different change detection models to identify deforested areas in the Amazon Rainforest, while investigating data processing and network training strategies that may improve detector effectiveness. Furthermore, although Transformer-based detectors are well-established in other domains, they remain underexplored for deforestation detection, justifying their incorporation and evaluation.

For deforestation identification through change detection, we hypothesize that self-attention mechanisms will benefit the models since deforested areas rarely appear in isolation due to logistical constraints. Purely convolutional models exhibit an inductive bias favoring local image features~\citep{swinunetrv2}; we therefore expect Transformer components to help mitigate this bias.

\section{Concepts}
\label{sec:concepts}

This section presents the theoretical foundations supporting the research, covering topics in Environmental Monitoring and Machine Learning.

\subsection{Land Use and Land Cover}

Mapping the Earth's emerged land areas is crucial for environmental monitoring, particularly for identifying changes that serve as key indicators of human interference in natural systems, which can cause drastic alterations to global ecology~\citep{luo2024decreased,qin2025impact,coulibaly2025effects,zhang2025biophysical,luc_impact}. In this context, the terms land use and land cover are commonly used and often interchangeable. However, they represent fundamentally different concepts: while land cover can be determined through simple observation, land use requires socioeconomic analysis of local activities~\citep{land_use_cover}.

Land cover refers to the physical material on the Earth's surface, that is, the surface material that interacts with electromagnetic radiation and is observed through aerial photography or satellite sensors. Land use, on the other hand, describes how people utilize the land; two common general classifications are the division between urban and agricultural use~\citep{land_use_cover}.

\subsection{Remote Sensing}
\label{subsec:remote_sensing}

In environmental monitoring, remote sensing refers to technologies measuring electromagnetic energy emanating from Earth's surface features, including land, oceans, and atmosphere. The properties of emitted or reflected electromagnetic waves enable identification and delineation of surface elements. When conducted via satellites, these technologies allow regular data capture from the same location, facilitating land use and cover change mapping~\citep{remote_sensing_book}.

Satellite sensor imagery is subject to various distortions, including sensor artifacts, solar conditions, atmospheric interference, and topographic variations~\citep{landsat_preprocessing}. Mitigation approaches include: (i) geometric corrections addressing distortions from sensor motion, speed variations, and satellite altitude; (ii) atmospheric corrections for effects caused by suspended particles and other atmospheric materials; and (iii) radiometric corrections involving noise treatment and reduction of Sun-Earth distance variation effects~\citep{remote_sensing_book}.

Remote sensing data undergo preprocessing before scientific distribution. These processes follow hierarchical levels, where each subsequent level involves greater processing. This project utilizes Landsat Level 2 data, providing surface reflectance values after atmospheric, geometric, and radiometric corrections~\citep{remote_sensing_models}.

\subsection{PRODES Methodology}
\label{sec:prodes_method}

The Project for Monitoring Deforestation in the Legal Amazon (PRODES) employs a standardized methodology for identifying deforested areas, as detailed in~\citep{prodes_method}. Since 1988, PRODES has systematically monitored primary forest loss across Brazil's Legal Amazon (ALB)~\citep{prodes_method}. According to its methodology documentation~\citep{prodes_method}, the project uses Landsat-compatible imagery, requiring 229 images to cover the entire ALB. Compatible satellites include Landsat-8/9, SENTINEL-2, and CBERS-4/4A, though Landsat imagery remains the primary data source for decision-making. Since 2016, PRODES has also provided annual deforestation maps for the Cerrado biome~\citep{cd_cerrado}.

The project maps deforestation polygons larger than 1 hectare but publishes only those exceeding 6.25 hectares for data consistency; it also provides polygons of areas obscured by clouds or their shadows~\citep{prodes_method}. This publication threshold creates a database limitation: while deforested areas may be clearly visible in a given year, they might not be published until reaching the 6.25 hectare threshold, potentially causing sudden appearance of accumulated deforestation when minor new clearing pushes previously unpublished areas past the 6.25 threshold.

\subsection{Tested Neural Network Architectures}
\label{subsec:architectures}

Our evaluation compared multiple neural network architectures for deforestation detection. This subsection briefly describes each approach.

\subsubsection{UNet\texttt{++}}
\label{sec:unet++}

Our architecture evaluation included UNet\texttt{++}, initially proposed by~\citet{unet++} for medical image segmentation. This network improves upon standard UNet~\citep{unet} by progressively enriching encoder feature maps before decoder integration, rather than using conventional skip connections.

The UNet\texttt{++} architecture builds upon a standard UNet framework but replaces simple skip connections with an intricate multi-path structure. In this design, feature maps from the encoder to decoder undergo convolutional operations that aggregate and process data from multiple layers, including information from different decoder levels.

\subsubsection{MultiResUNet}
\label{sec:multiresunet}

Another convolutional neural network architecture evaluated in this study is the MultiResUNet, proposed by~\citet{multiresunet} as an evolution of the conventional UNet for medical image segmentation.

This architecture aims to improve UNet's performance by addressing two main challenges: (i) the scale variation of objects within images, and (ii) the semantic discrepancy between encoder and decoder levels connected by skip connections, a problem that UNet\texttt{++} also addresses, with a similar solution as discussed in Subsection~\ref{sec:unet++}.

To handle challenge (i) of scale variation, MultiResUNet employs so-called MultiRes blocks, which are based on three fundamental concepts: (1) the use of filters of different dimensions to capture features at multiple scales, inspired by the Inception architecture~\citep{inception}; (2) the property that cascaded convolutions can have the same receptive field as larger filters (two 3$\times$3 convolutions are equivalent to one 5$\times$5, and three 3$\times$3 convolutions are equivalent to one 7$\times$7), as demonstrated by~\citet{rethinking_inception}; and (3) the inclusion of skip connections following the approach of residual blocks~\citep{resnet}. In practice, each MultiRes block consists of three sequential 3$\times$3 convolutional layers, whose outputs are concatenated to simulate the effect of different filter sizes. Moreover, the block's input passes through a 1$\times$1 convolution and is added to the concatenated outputs, following the residual learning principle.

To address challenge (ii) of semantic discrepancy, the architecture introduces Res Paths (residual paths), which replace conventional skip connections. These paths consist of simple residual blocks, composed of 3$\times$3 convolutions interspersed with skip connections that use 1$\times$1 convolutions.

\subsubsection{TransUNet}
\label{sec:transunet}

The TransUNet, proposed in 2021 by~\citet{transunet_new}, was one of the first models to integrate Transformer architectures for medical image analysis. This model maintains a structure analogous to the traditional UNet, but incorporates both convolutional encoder-decoder components and Transformer layers.

The model uses a convolutional encoder for initial feature extraction, with a corresponding convolutional decoder , connected via skip connections. Between these two components, it inserts a Transformer encoder that receives the flattened feature map, divides it into patches, and encodes them for processing by a Transformer encoder~\citep{transunet_ori}.

\subsubsection{SwinUNETR-V2}
\label{sec:swinunetr-v2}

While convolutional operations introduce an inductive bias of locality, which may limit model performance, traditional Transformers eliminate it, but makes efficient model training challenging~\citep{swinunetrv2}. To address this fundamental limitation, among other issues, \citet{swin_transformer} developed the Swin Transformer, an architectural variation that incorporates cyclic sliding windows into the self-attention mechanism, thereby reintroducing locality in a controlled manner while preserving the benefits of global attention.

The SwinUNETR-V2 (2023)~\citep{swinunetrv2} is a medical image segmentation model that adapts the basic UNet structure by incorporating Swin Transformers. At each encoder stage, the architecture employs a residual convolutional block followed by two Swin Transformer blocks, which utilize the cyclic sliding window approach, and a patch merging block that combines neighboring patches and reduces their dimensionality through a linear layer. The decoder maintains a convolutional structure connected to the encoder via skip connections, with the addition of residual convolutional blocks in these pathways.

\section{Related Work}
\label{sec:related_work}

This section discusses relevant studies related to our research topic. Our work focuses on identifying deforestation areas in the Amazon Rainforest using change detection models. We utilize annual deforestation data from the PRODES Project combined with corresponding satellite imagery of the region and time period. The related work covers three main aspects: (a) generation of appropriate datasets, (b) potential approaches for creating change detectors, and (c) existing change detection models in the literature. Within this last aspect, we examine both architectures previously tested for deforestation detection and models that, to our knowledge, have not been specifically applied to this domain.

Regarding aspect (a), the work proposed by~\citet{pozzobon20} employed images from three scenes of different regions captured by the Landsat 8 satellite across multiple years. One region served as the test set, while the other two were used for training. For our preliminary tests, we adopted this same strategy, even utilizing the same Landsat scenes. However, \citeauthor{pozzobon20} manually refined the deforestation masks provided by PRODES, a step that could not be replicated or automated in our tests.

Another proposal that extensively addressed dataset generation was the work by~\citet{maretto21}. Their approach described a systematic procedure for selecting Landsat images while minimizing forest occlusion by clouds, including the combination of multi-date imagery to ensure complete area coverage. However, as the authors themselves acknowledged, this method may introduce artifacts in the final composite image. Notably, in addition to the change map identifying deforested areas, they created a cloud and shadow mask that overlays the change map for occluded pixels, thereby reducing model susceptibility to such noise.

The works of~\citet{cd_cerrado} and~\citet{cd_amazon_comparison} enhance the dataset images through the addition of supplementary spectral bands. Both studies employ the Normalized Difference Vegetation Index (NDVI), which is derived from simple arithmetic operations (addition, subtraction, and division) applied to bands 4 and 5 of the Landsat-8 OLI sensor. This derived band effectively highlights vegetation areas and can identify regions of vegetation stress~\citep{ndvi}.

Regarding aspect (b) of the analysis, various techniques exist for generating change maps, with deep learning-based approaches becoming the primary research focus~\citep{cd_review_2022}. These methods can be categorized into: (1) fully supervised approaches when annotated change maps are available; (2) unsupervised methods when working with unannotated data; and (3) transfer learning-based techniques that leverage knowledge from related problems. For our study, we employ fully supervised methods since annotated ground truth data can be generated for this application.

Furthermore, we can categorize the methods based on their network architecture. \citet{cd_review_2020} classify these approaches into: (i) single-stream, (ii) double-stream, and (iii) multi-model architectures. The multi-model methods (iii) integrate different models into a hybrid framework, with training potentially occurring at different stages. The distinction between single-stream (i) and double-stream (ii) methods lies in their data flow organization.

\begin{figure}[!htb]
\centering
\subfloat[single-stream network]{\includegraphics[width=0.99\linewidth]{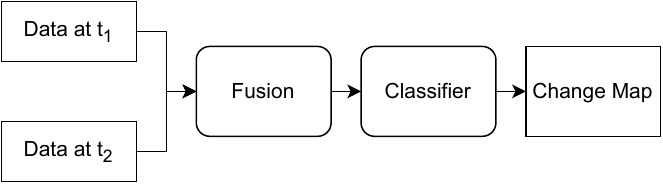}} \\
\subfloat[double-stream network]{\includegraphics[width=0.99\linewidth]{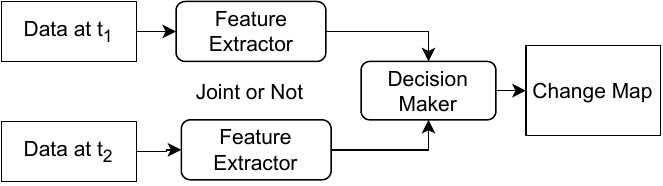}}
\caption{Diagram of (a) single-stream and (b) double-stream networks for change detection. Adapted from~\citep{cd_review_2020}.}
\label{fig:cd-methods}
\end{figure}

In single-stream approaches, data passes through the network only once to generate the change map. A representative example is the Early Fusion (EF) architecture, where temporal data at $t_1$ and $t_2$ are concatenated before being fed into the model. Conversely, double-stream methods (ii) maintain separate data flows for $t_1$ and $t_2$ within the network, combining them only at later stages to produce the change map. Siamese networks are a representative example of this architecture~\citep{cd_review_2020}. Figure~\ref{fig:cd-methods} illustrates these architectural differences, particularly showing how temporal data is combined at different processing stages in each approach.

Regarding aspect (c), \citet{maretto21} proposed two modified versions of UNet~\citep{unet}, a popular convolutional neural network for segmentation, where one version employs an Early Fusion strategy to combine temporal data while the second combines data after feature extraction, the latter showing slightly superior effectiveness despite higher implementation complexity and computational costs. \citet{pozzobon20} also implemented various convolutional neural network variants and compared them with classical algorithms, consistently using an EF strategy that can be adapted from semantic segmentation models, making them promising candidates for our tests. Similarly, \citet{cd_cerrado} utilized UNet combined with a recurrent neural network (RNN), where the RNN first processes temporal satellite data to generate deforestation probability maps that feed into UNet alongside terrain slope data, ultimately producing the final deforestation classification.

In addition to studies focused specifically on deforestation identification, other untested models remain relevant to this project, particularly those employing Transformer self-attention mechanisms. \citet{rs_transformer} proposed a model using a convolutional backbone (ResNet-18~\citep{resnet}) for feature extraction, refined through a bitemporal image \textit{Transformer}, with final classification performed by a convolutional classifier to generate the change map. This approach demonstrated superior performance compared to purely convolutional models, including UNet itself.

\citet{change_former} proposed another Transformer-based model for change detection applications, the ChangeFormer. This siamese network consists of a series of blocks that reduce data resolution and feed it to a Transformer. The outputs from each of these blocks for images at $t_1$ and $t_2$ are then compared, combined, processed through a resolution enhancement stage, and finally passed to a convolutional classifier that generates the change map. The model's effectiveness surpasses both purely convolutional approaches and other self-attention-based methods.

Regarding ChangeFormer, but discussing aspects of (c), \citet{amazon_changeformer} evaluated this model, originally developed for urban change detection, for deforestation monitoring using Sentinel-2 satellite imagery. While they used different imagery sources than Landsat (employed in our study), their reference masks were generated from PRODES data, similar to our approach. Beyond NDVI, their work incorporated other spectral band combinations during training: Color-Shifted Infrared and the Enhanced Vegetation Index (EVI). The authors implemented percentile-based data normalization, which we used to generate the figures in this paper, but not for training. Additionally, they applied a training data filtering process, manually selecting only high-quality rasterizations and excluding cases with less than 10\% deforestation change, aiming to reduce dataset imbalance.

\section{Methodology}

This section presents the methodology developed for this study to address the research objectives. The section is divided into subsections describing the data acquisition and processing pipeline to combine and transform multi-source data into a unified dataset suitable for our problem, and the training and validation procedures, including preprocessing, postprocessing, and model ensemble techniques.

\subsection{Data Collection and Preparation}
\label{sec:data_prep}

\begin{figure}[!htb]
\centering
\includegraphics[width=1\linewidth]{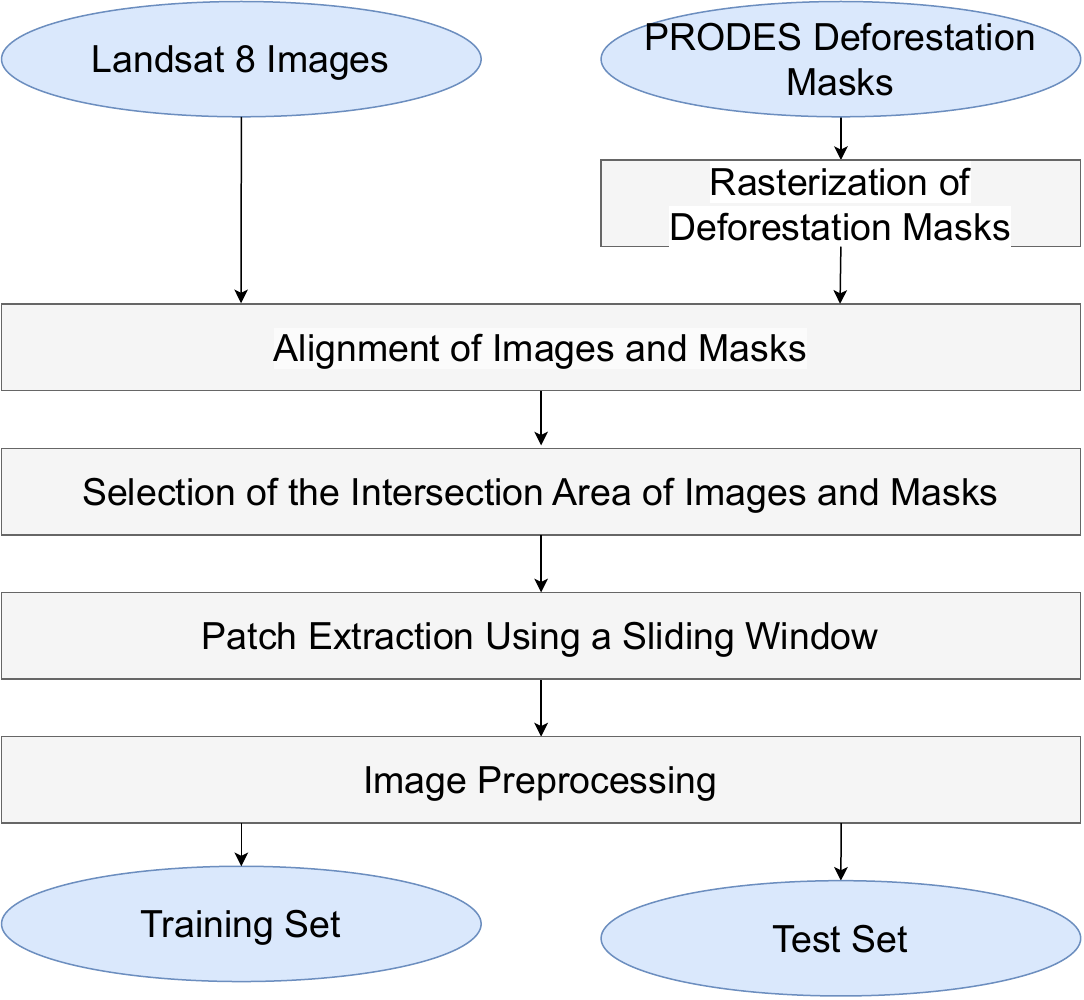}
\caption{Methodological workflow for data processing, from PRODES and Landsat sources to the final training and test sets for change detection models.}
\label{fig-method:data_process}
\end{figure}

This subsection describes the dataset generation process, illustrated in Figure~\ref{fig-method:data_process}. We divide the procedure into two main stages: (i) data collection, specifically satellite imagery and deforestation masks; and (ii) creation of the change detection dataset, which involves combining information from both sources to produce a coherent unified dataset, along with image preprocessing details. We present the image transformations applied to the combined dataset to facilitate feature extraction by the models.

\subsubsection{Data Acquisition}
\label{subsec:acquiring_data}

The data used for training and validating change detection models comes from two primary sources: USGS~\citep{usgs} and PRODES~\citep{prodes}. The former provides remote sensing imagery without annotations of deforested regions, while PRODES supplies annual deforestation data that enables construction of a suitable dataset for our target problem.

PRODES data can be obtained through the TerraBrasilis platform\footnote{\url{http://terrabrasilis.dpi.inpe.br/}}. Among available project data, we use the ``Annual Deforestation Increment'', a vector shapefile~\citep{shapefile} containing post-2008 yearly deforestation records. The deforested regions in this file contain class labels formatted as dYYYY, where YYYY represents the detection year when compared with the previous period.

\begin{figure}[!htb]
\subfloat[Landsat scene locations]{\includegraphics[width=\linewidth]{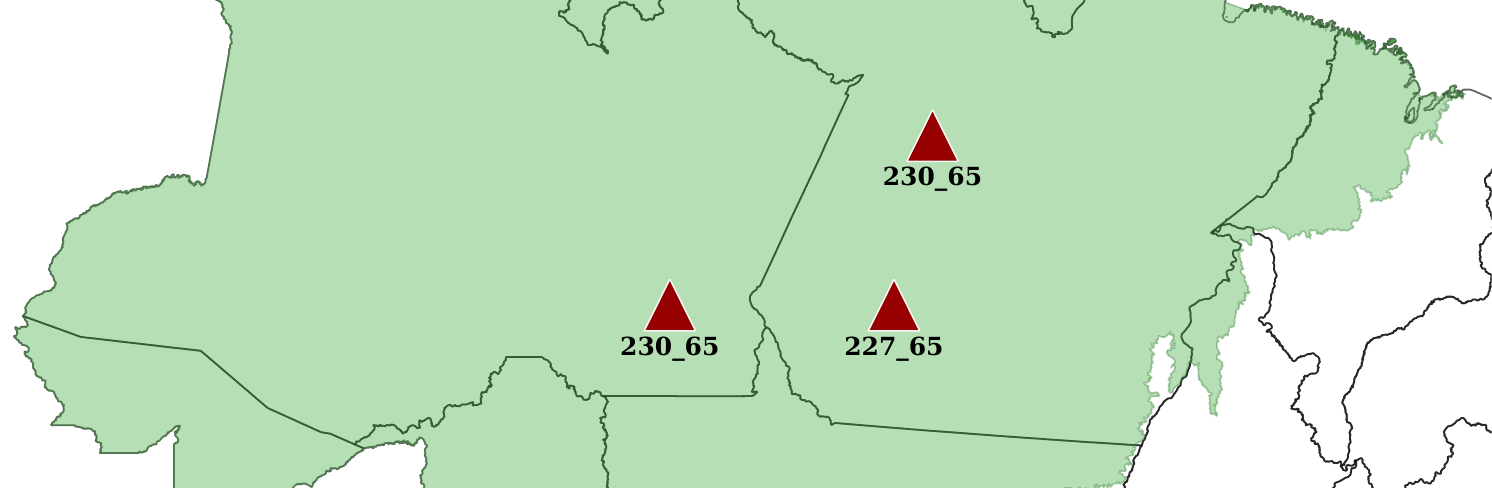}} \\
\subfloat[227\_63]{\includegraphics[width=0.3\linewidth]{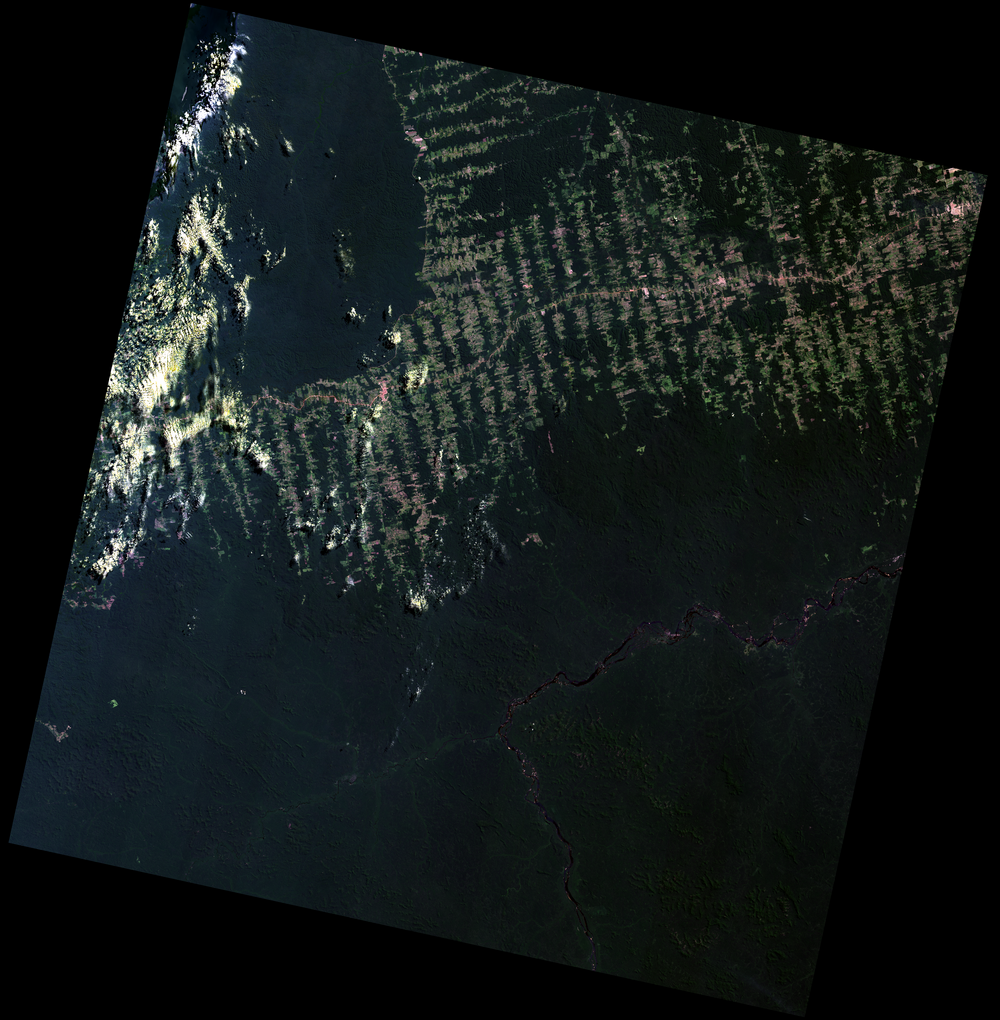}} \hfill
\subfloat[227\_65]{\includegraphics[width=0.3\linewidth]{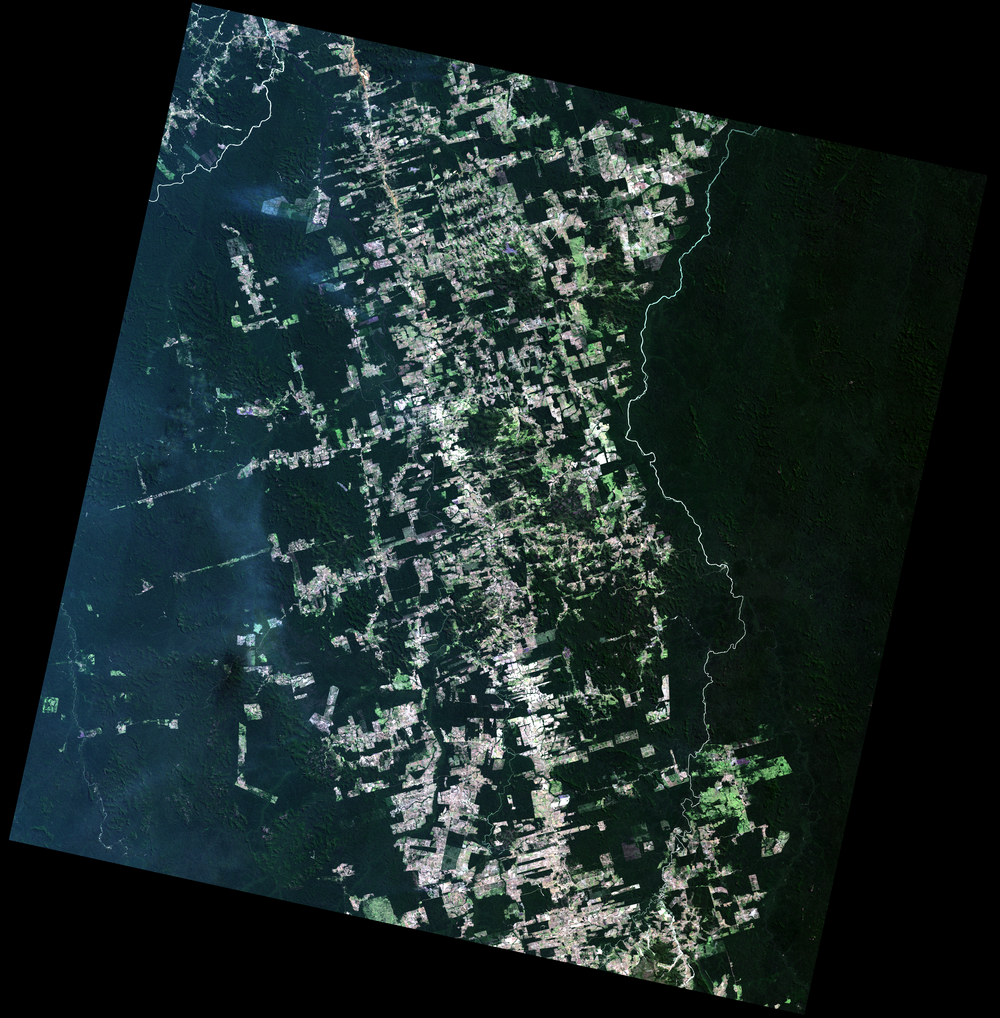}} \hfill
\subfloat[230\_65]{\includegraphics[width=0.3\linewidth]{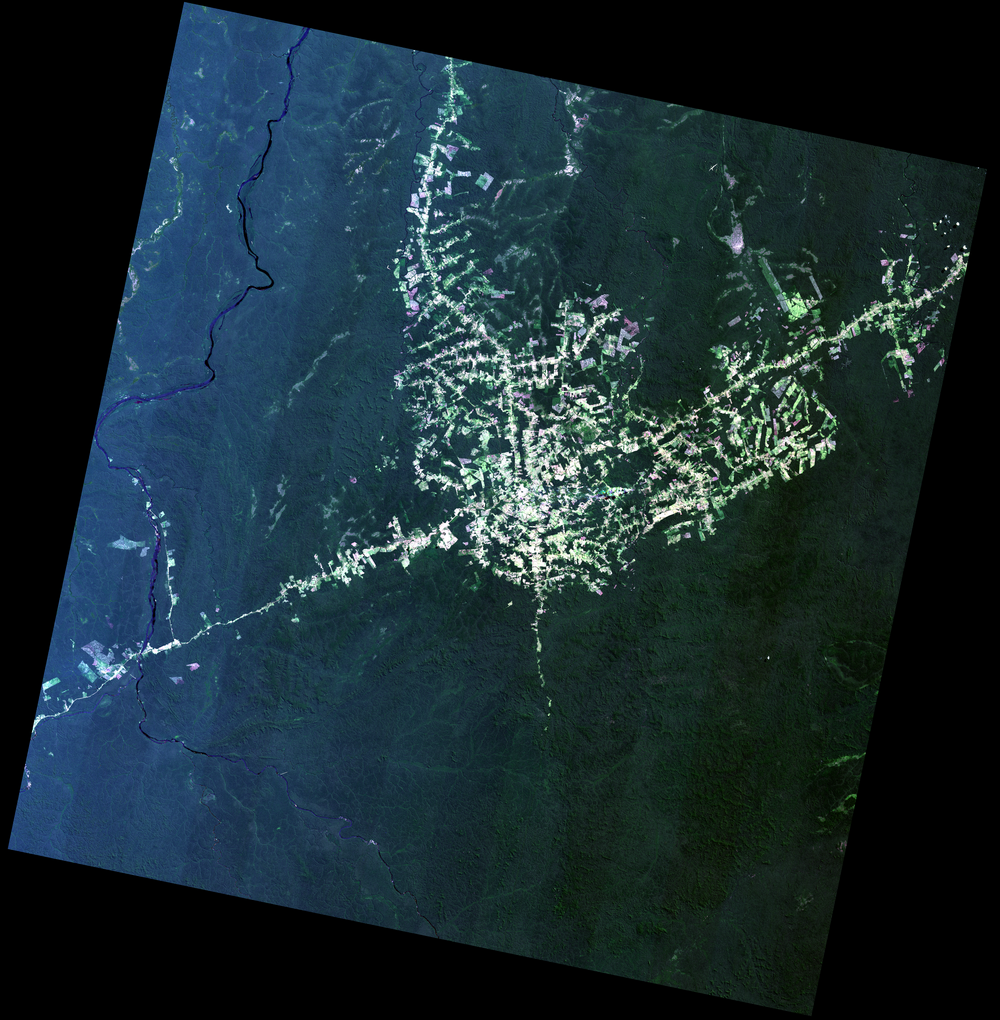}}
\caption{(a) Map showing locations of selected Landsat scenes for training and testing. (b-d) 2019 imagery of Landsat scenes. Adapted from~\citep{pozzobon20}.}
\label{fig:scenes}
\end{figure}

Following the approach proposed by~\citet{pozzobon20}, we selected satellite imagery from three Landsat scenes (Table~\ref{tab:data_origin} and Figure~\ref{fig:scenes}) near the Trans-Amazonian Highway (BR-230) and Cuiabá-Santarém Highway (BR-163). We used measurements from the Landsat 8 OLI sensor, available through Collection 2 Level-2 Landsat products.

\begin{table}[!htb]
\centering
\small
\setlength{\tabcolsep}{3.0mm}
\caption{Landsat scenes and acquisition dates of remote sensing imagery used for creating the change detection dataset.}
\label{tab:data_origin}
\begin{tabular}{lccc}
\toprule
\multirow{2}{*}{\textbf{Landsat Scene}} & \multicolumn{3}{c}{\textbf{Acquisition Date}} \\ \cmidrule(lr){2-4}
& 2017 & 2018 & 2019 \\
\midrule
227\_63 & 07/18 & 07/21 & 07/24 \\
227\_65 & 07/18 & 07/21 & 07/24 \\
230\_65 & 07/23 & 06/24 & 07/13 \\
\bottomrule
\end{tabular}
\end{table}

\subsubsection{Creation of the Change Detection Database}
\label{sub:data_merge}


This subsection describes the procedures performed to process the data obtained by the previous methods and transform them into a database suitable for training a change detector. For this purpose, we define $I_k^{y}$ as a band of the image $I^{y}$ from a Landsat scene, where $k \in \left[ 1,7 \right]$ represents one of the channels measured by the OLI instrument we used; and $y \in \{1,2\}$ indicates whether the image is the oldest, $y=1$, or the most recent, $y=2$, among the two collected for the region of interest. These 7 channels were chosen because they have the same resolution of 30 m. Although band 9 has the same resolution, its primary utility is to detect Cirrus clouds composed of small ice crystals~\citep{landsat_manual}, which was considered redundant with the quality assessment band that already provides information about their presence, and was therefore discarded.

Additionally, we define $M$ as the set of annual deforestation regions acquired from PRODES that are part of the selected Landsat scenes; and $M^{1-2}$ as the subset composed only of regions deforested between the years corresponding to $I^1$ and $I^2$. This mask is provided in a vector format, so to be compared with the outputs of the change detection models, it must undergo a rasterization process.

The steps to generate the processed database can be divided and organized according to the diagram in Figure~\ref{fig-method:data_process}, whose description is:

\begin{enumerate}

\item \textbf{Rasterization of Deforestation Masks}: the deforestation regions detected between the dates of the two images are rasterized by filling polygons in a raster image with the same resolution as $I^y$. Pixels corresponding to deforested regions receive the value 1, while the rest are set to zero. The rasterized mask will be denoted as $M_r^{1-2}$.
	
\item \textbf{Alignment of Images ($I^1$, $I^2$) and Masks ($M_r^{1-2}$)}: although the Landsat scenes correspond to the same location, there is a slight variation in image position from one year compared to another. Additionally, the deforestation mask and Landsat scenes use different coordinate references, thus a reprojection of $M_r^{1-2}$ to match the Landsat images is necessary. The three matrices must be aligned so that each point in one image has its equivalent in the other two, representing the same planetary position within resolution limits. The alignment was performed automatically through affine transformations in the real coordinate space of the images.
	
\item \textbf{Selection of the Intersection Area of Images and Masks}: after aligning the scenes and the mask, the intersection of the three images is computed and extracted, resulting in three images of the same dimensions where each pixel $(i,j)$ represents the same location on the globe for all of them. We denote the resulting matrices from this cropping as $I^{'1}$, $I^{'2}$, and $M_r^{'1-2}$.
	
\item \textbf{Patch Extraction Using a Sliding Window}: the two Landsat images and the mask are cropped using a sliding window technique, defining $i_{j}^{'y} \in I^{'y}$ and $m_{r,j}^{'1-2} \in M_r^{'1-2}$, $j \in \left\{1,\dots,n\right\}$, where $n$ is the number of images generated in this step. For this purpose, all matrices are concatenated so that a single pass of the algorithm applies to all data, maintaining alignment. This window has a height $h_w$ and width $w_w$ of 256 pixels, with a stride of 200 pixels. As can be seen in Figure~\ref{fig:scenes}, the images contain dark regions on the sides due to scene rotation, which lack data and are not suitable for training. Therefore, in cases where the proportion of null pixels in $i_j^{'1}$ and $i_j^{'2}$ combined exceeds 5\%, the sample is discarded.
	
\item \textbf{Image Preprocessing}: The pixels of the Landsat images are provided in 16 bits per band. However, its intensities are not well distributed. To address this, the images undergo a normalization process, as described in Equation~\ref{eq:standardize}. Afterwards, we tested applying histogram equalization to the images and evaluated its impact (Subsection~\ref{sec:method-eq-hist}):
\begin{equation}
i_{j,k}^{''y} = \frac{\left( i_{j,k}^{'y} - \text{avg}(I_k^{'y}) \right)}{\text{std}(I_k^{'y})}\raisepunct{,}
\label{eq:standardize}
\end{equation}
\noindent where $\text{avg}(X)$ represents the mean of the values in $X$, $\text{std}(X)$ is the standard deviation of $X$, and $j \in \{ 1,\dots, n \}$.
	
Additionally, an extra channel is included in each image to store the Normalized Difference Vegetation Index (NDVI), calculated as shown in Equation~\ref{eq:ndvi}.
\begin{equation}
\text{NDVI} = \frac{B2 - B1}{B2 + B1}\raisepunct{,}
\label{eq:ndvi}
\end{equation}
\noindent where $B1$ is the reflectance value of the red visible band and $B2$ is the reflectance value of the near-infrared band.

\item \textbf{Sample Storage}: each set $i_j^{''1}$, $i_j^{''2}$, and $m_{r,j}^{'1-2}$, $j \in \left\{1,\dots,n\right\}$ is saved in a lossless compressed file, organized by scene, year, and whether it belongs to the training or test set. More specifically, the data related to scene 230\_65 compose the test set; the other two scenes are used for training, thus ensuring separation between these two sets, including the biome regions in which they are located.
	
\end{enumerate}

\begin{figure}[!htb]
\centering
\subfloat[2018]{\includegraphics[width=0.29\linewidth]{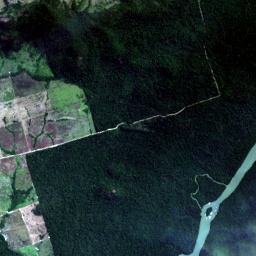}} \hspace{1em}
\subfloat[2019]{\includegraphics[width=0.29\linewidth]{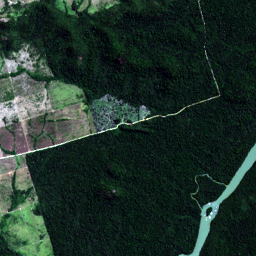}} \hspace{1em}
\subfloat[Mask]{\includegraphics[width=0.29\linewidth]{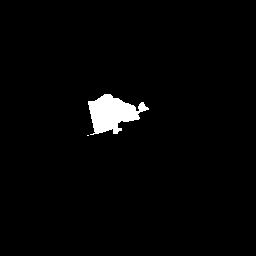}} \\
\subfloat[2018]{\includegraphics[width=0.29\linewidth]{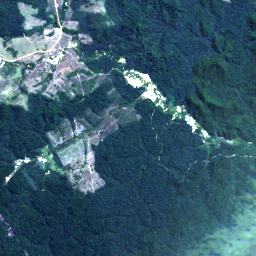}} \hspace{1em}
\subfloat[2019]{\includegraphics[width=0.29\linewidth]{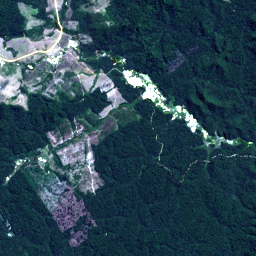}} \hspace{1em}
\subfloat[Mask]{\includegraphics[width=0.29\linewidth]{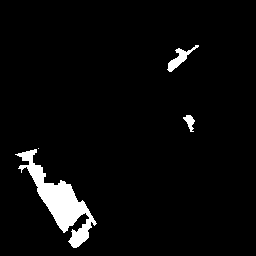}}
\caption{Examples of samples from the dataset adapted for the problem of identifying deforested areas through change detection. The images were adjusted to be visualized in an appropriate color space.}
\label{fig-method:train_samples}
\end{figure}

At the end of this procedure, 5,665 samples were produced. Some of these examples can be visualized in Figure~\ref{fig-method:train_samples}.

\subsection{Training and Validation}
\label{sec:train_val}

\begin{figure}[!htb]
\centering
\includegraphics[width=\linewidth]{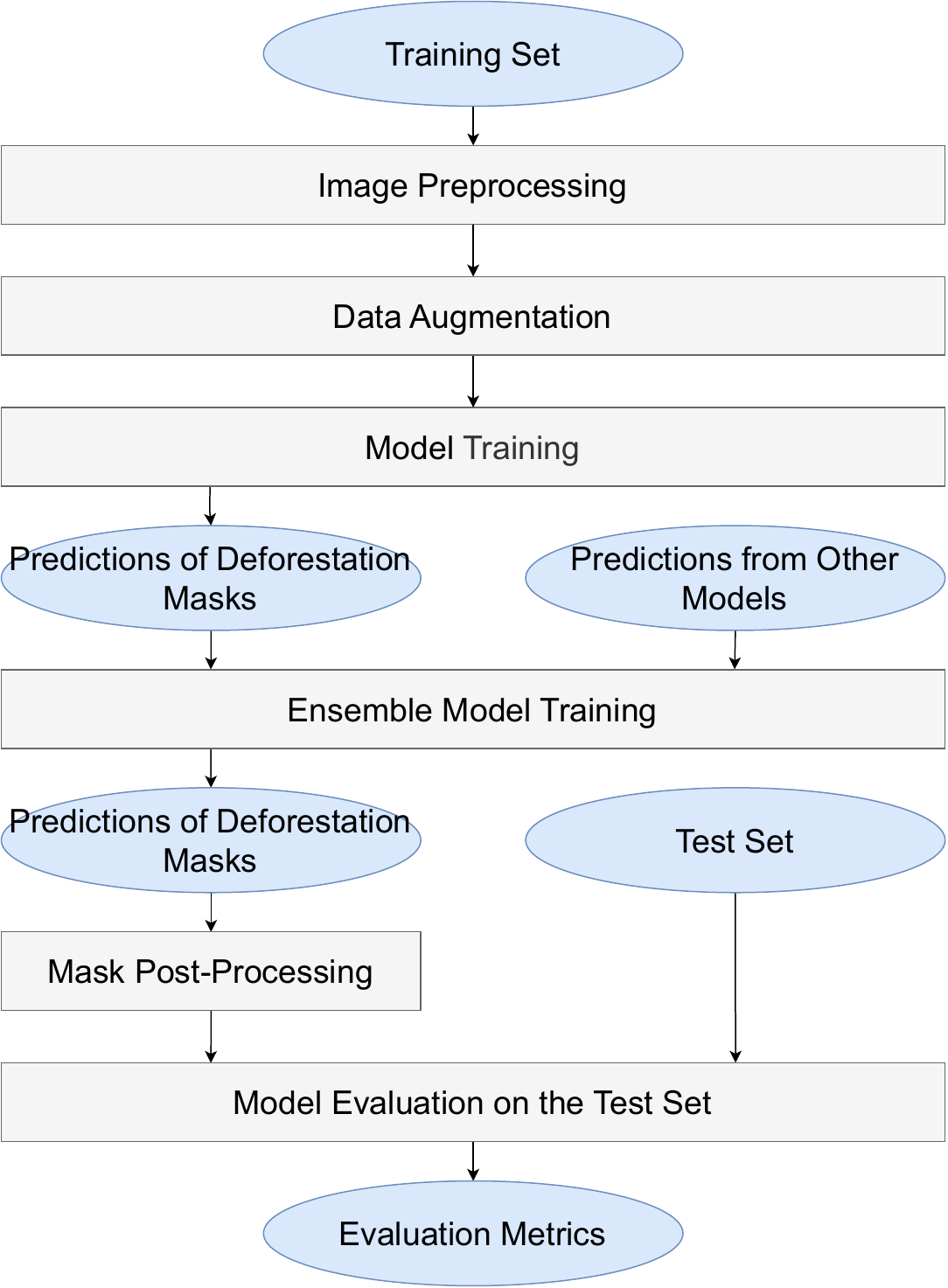}
\caption{Diagram of the training and validation methodology steps for deforestation detection models, including individual model training, prediction combination (ensemble) and performance metrics evaluation.}
\label{fig-method:train_val}
\end{figure}

In this subsection, we define the details of the model training procedure, their combination, and the adaptations applied to tailor them to the problem at hand. We also detail the evaluation metrics used to measure the effectiveness of the models tested on the problem and the validation protocol. The process flow is illustrated in Figure~\ref{fig-method:train_val}, which describes the steps of training the individual models, generating predictions, combining the models (ensemble), and the final evaluation of performance metrics. Note that the diagram includes an ``Image Preprocessing'' step that also appeared in Figure~\ref{fig-method:data_process}, but in practice, they involve different techniques. In the current case, the processing is performed during the initialization of the training models, not prior to saving the dataset to disk.

\subsubsection{Histogram Equalization}
\label{sec:method-eq-hist}

This subsection addresses the histogram equalization process~\citep{hist_eq}, which is part of the ``Image Preprocessing'' stage in Figure~\ref{fig-method:train_val}. The equalization is performed right before the training step, with the objective of increasing the contrast of the original image. This technique is particularly useful in the context of Landsat images, which, as previously discussed, often exhibit low-contrast intensity distributions. However, it should be noted that the technique was not applied in all tests, as it was incorporated into the workflow later; we will present in the results the effect of equalization on the addressed problem.

Since we are dealing with a $C$-channel image, each channel is treated independently, as if it were a monochromatic image. Furthermore, because we are using small patches, a global histogram equalization is sufficient to achieve the desired results, without the need for more complex techniques, such as local equalization.

To formalize the histogram equalization process, we begin by defining the probability of a pixel in the image having value $k$:
\begin{equation}
p_k = \frac{n_k}{N}\raisepunct{,}
\label{eq:prob_pixel}
\end{equation}
\noindent where $n_k$ is the number of pixels with intensity $k$, whereas $N$ is the total number of pixels in the image.

This probability $p_k$ represents the intensity distribution of the image and forms the basis for calculating the cumulative distribution function (CDF). The CDF is defined as the cumulative sum of probabilities $p_k$ for all intensity values up to $k$:
\begin{equation}
\text{CDF}(k) = \sum_{i=0}^{k} p_i,
\label{eq:cdf}
\end{equation}

The histogram equalization is then defined as the transformation of each intensity value $k$ into a new value $r_k$, given by the rounded normalized CDF:
\begin{equation}
r_k = \text{round} \left( (L - 1) \cdot \text{CDF}(k) \right),
\label{eq:equalizacao}
\end{equation}
\noindent where $r_k$ is the equalized value corresponding to the original intensity $k$ and $L$ is the total number of possible intensity levels (for example, 256 for 8-bit images).

\subsubsection{Replacement of Clear-Cut Deforestation Patterns with Fire Use}
\label{sec:magenta_method}

\begin{figure}[!htb]
\centering
\subfloat{\includegraphics[width=0.32\linewidth]{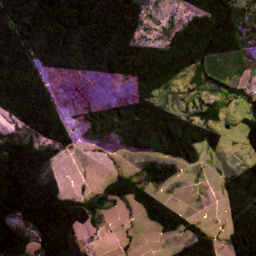}} \hfill
\subfloat{\includegraphics[width=0.32\linewidth]{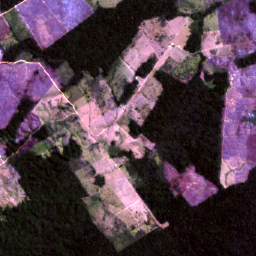}} \hfill
\subfloat{\includegraphics[width=0.32\linewidth]{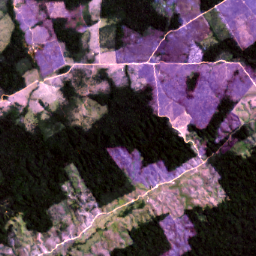}}
\caption{Examples of magenta regions detected in the dataset. The images show the characteristic spectral signature of this pattern, which often leads to misclassification by change detection models.}
\label{fig:magenta_patterns}
\end{figure}

This subsection addresses the removal process of magenta regions (Figure~\ref{fig:magenta_patterns}), which constitutes part of the ``Image Pre-processing'' stage in Figure~\ref{fig-method:train_val}. The removal is performed during dataset loading, though not all tests employed this technique. Below, we discuss the impact of this procedure on model effectiveness.

The dark magenta region is, in most cases, associated with a clear-cut deforestation pattern with fire use~\citep{prodes_method}. This pattern is characterized by areas where vegetation was completely removed and affected by fire, resulting in a distinct spectral signature. The change detection models demonstrated difficulty in correctly handling this pattern, often misclassifying it.

To define the magenta regions, we conducted a statistical analysis of the spectral bands in the training set images. We calculated the mean and standard deviation of each band for a set of pixels defined as magenta, allowing us to identify regions whose spectral values differed significantly from the rest of the scene. The selected region for analysis was:

\begin{itemize}
	
\item \textbf{Scene 227\_063 - 2019}:

\begin{itemize}

\item (4266, 3948) - (4292, 3969).

\end{itemize}

\end{itemize}

The listed coordinates represent the bounding box selected to define the region of interest within the Landsat images. From this region, we calculated the mean and standard deviation of the spectral bands to characterize the spectral signature of magenta pixels.

A pixel $x$ is classified as magenta if its spectral value in each band is within the range of the mean $\pm$ 1.2 standard deviations:
\begin{equation}
\mu_b - 1.2\sigma_b \leq p_b \leq \mu_b + 1.2\sigma_b, \quad \forall b \in B,
\label{eq:magenta_threshold}
\end{equation}
\noindent where $p_b$ represents the spectral intensity of the pixel in band $b$, $\mu_b$ is the mean intensity of magenta pixels in band $b$, $\sigma_b$ is the standard deviation of magenta pixels in band $b$, amd $B$ represents the set of analyzed spectral bands.

Pixels that meet the criterion of Equation \eqref{eq:magenta_threshold} are considered to belong to the magenta region and, therefore, can be removed during the pre-processing step by replacing them with a more common deforestation pattern. For this purpose, we define a texture bounding box extracted from scene 227\_063 of 2019, with coordinates (5926, 2126) - (5973, 2151). This region was chosen because it represents a typical spectral pattern of deforestation, without the problematic magenta pattern. From this bounding box, we created a repeated texture image with the same dimensions as the original image. For each detected magenta pixel, we replaced its value with the corresponding pixel in the texture image.

\begin{figure}[!htb]
\centering
\subfloat[Before removal]{\includegraphics[width=0.49\linewidth]{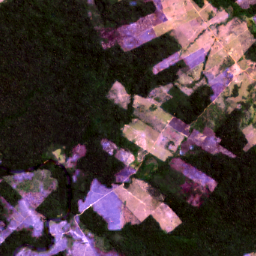}} \hfill
\subfloat[After removal]{\includegraphics[width=0.49\linewidth]{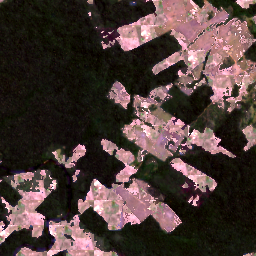}}
\caption{Comparison between the original image and the image after the removal of magenta regions. Replacing the magenta areas with a common deforestation texture pattern results in a more consistent image for analysis.}
\label{fig:magenta_removal}
\end{figure}

Figure~\ref{fig:magenta_removal} illustrates the effect of removing magenta regions. The figure shows the original image (left) and the image after replacing the magenta areas with the texture pattern (right). Note that removing these regions results in a more homogeneous image, which can improve the accuracy of change detection models.

\subsubsection{Model Training and Selection}
\label{sec:train-model}

In this subsection, we address the process of training change detection models and selecting the best one using the validation set. Although each experiment has particularities in its execution, some decisions were common to all cases and will be detailed here. Specific variations for each test will be described along with the presentation of the respective experiments.

The training procedure adopted follows a typical deep learning workflow, consisting of a loop that iterates over the dataset, applies stochastic transformations to the images (data augmentation), and feeds them to the model. The model's predictions are then compared to the ground truth deforestation masks using a loss function, and the model weights are updated according to the computed gradient.

For optimization, we used the Adam algorithm~\citep{adam} with parameters $\beta_1 = 0.9$, $\beta_2 = 0.999$, $\epsilon = 10^{-8}$, and a learning rate of $10^{-4}$, except for the MultiResUNet model, for which $3\cdot 10^{-4}$ was used. In addition, we applied L2 regularization with a coefficient of $10^{-4}$ to prevent overfitting. Preliminary tests with learning rate schedulers were performed, but the observed effects were negligible, so we opted to keep a fixed learning rate for simplicity. Training was carried out with a batch size of 32.

For the loss function, we adopted the focal loss~\citep{focal_loss}, which is especially useful in problems with imbalanced classes, such as our case of change detection in images. The focal loss equation is defined as:
\begin{equation}
\mathcal{L} = -\alpha_t (1 - p_t)^\gamma \log(p_t),
\end{equation}
\noindent where:
\begin{itemize}
\item $\mathcal{L}$: loss function value.
\item $\alpha_t$: balancing factor between the positive and negative classes, defined as:
\begin{equation*}
\alpha_t = \begin{cases}
\alpha & \text{if the true class is 1}, \\
1 - \alpha & \text{if the true class is 0},
\end{cases}
\end{equation*}
\noindent where the value $\alpha \in \left[ 0,1 \right]$ corresponds to the weight assigned to the positive class. We used $\alpha = 0.25$.
\item $p_t$: model probability for the correct class. For a binary class problem, where class 1 is the positive class, $p_t$ is defined as:
\begin{equation*}
p_t = \begin{cases}
p & \text{if the true class is 1}, \\
1 - p & \text{if the true class is 0}.
\end{cases}
\end{equation*}
Here, $p$ is the model's probability of correctly classifying class 1 (positive class).
\item $\gamma$: focusing parameter, which controls the weight assigned to hard examples; larger $\gamma$ values increase the focus on harder examples. We used $\gamma = 2$.
\end{itemize}

The choice of $\alpha = 0.25$ may seem counterintuitive, since the rarest class in the problem is the positive class, so values above 0.5 would be expected. However, as pointed out by the authors~\citep{focal_loss}, the choice of $\gamma$ tends to have a greater impact on the final result, and increasing $\gamma$ while reducing $\alpha$ often yields the best results.

In addition to focal loss, we tested other loss functions, such as weighted cross-entropy, focal tversky loss~\citep{focal_tversky}, combo loss~\citep{combo_loss}, and asymmetric focal loss~\citep{asymmetrical_focal_loss}. However, only focal tversky loss produced results comparable to those obtained with focal loss, which proved to be the best fit for our problem.

We allocated 20\% of the training set for validation, allowing continuous monitoring of the model's performance. Training was conducted for a total of 400 epochs, and in the end, we selected the model corresponding to the epoch with the best F1-score on the validation set.

Additionally, another parameter we selected on the validation set is the optimal threshold for binarizing the output. That is, the model's output, which ranges from 0 to 1 due to the sigmoid function, is converted into a binary result through a threshold, where any pixel strictly greater than it is classified as the positive class; otherwise, it is classified as negative. At the end of training, the best model is evaluated on the validation set using different thresholds, and we select the one that yields the best F1-score.

\subsubsection{Data Augmentation}
\label{sec:data-aug}

In this subsection, we describe the data augmentation techniques used during model training. These techniques follow a stochastic approach, in which random transformations are applied to the input images at runtime. The use of data augmentation is essential in this work, since the size of the training set is limited.

The data augmentation techniques applied consist of three main approaches, which are applied sequentially to both images at the same time. Thus, the occurrence of one technique does not prevent the occurrence of the others, further increasing the diversity of transformations:

\begin{enumerate}
	
\item \textbf{Random crop with resizing}: A random crop is performed on the image, followed by resizing to restore the original dimensions. The aspect ratio of the image is preserved, and the crop scale varies between 0.4 and 1, ensuring that the resulting image retains a significant proportion of the original content.
	
\item \textbf{Vertical flip}: A vertical flip is applied to the image with a probability of 50\%.
	
\item \textbf{Horizontal flip}: Similar to the vertical flip, a horizontal flip is applied with a probability of 50\%.
	
\end{enumerate}

These techniques increase the diversity of the dataset, exposing the model to different variations of the original images and improving its generalization ability.

\subsection{Post-Processing of Deforestation Masks}
\label{sec:mask_post_processing}

In this subsection, we address the post-processing applied to the deforestation masks predicted by the model. As discussed in Section~\ref{sec:prodes_method}, PRODES only publishes deforestation regions with a minimum area of 6.25 hectares, although smaller areas are also considered in its methodology. The prediction masks operate at a spatial resolution where each pixel represents 30 meters of the Earth's surface. Equating the mask area to the minimum area reported by PRODES, a deforestation of 6.25 hectares corresponds to approximately 70 pixels.

In practice, considering that the model may make errors and that there are approximations arising from the rasterization process, we decided to adopt a more conservative approach. To this end, we performed connected component detection on the predicted mask, using 4-connectivity. Based on this analysis and following the more conservative approach, we removed incremental deforestation regions with up to 50 pixels, ensuring that only significant areas are retained in the final masks.

A limitation of this method is related to the evaluation of the deforested region using patches. It is possible that, in dividing the patches, an incremental deforestation region may be fragmented, causing parts of it to meet the small-area removal criterion undesirably. This undesired removal may increase the number of false negatives, reducing the model's sensitivity. Although it would be possible to modify the ground truth mask to avoid this issue, we chose not to do so, as such alteration could contaminate the validation between models and approaches with and without the processing.

\subsection{Evaluation Metrics}

To assess the effectiveness of the models to be tested, we compare the outputs of the networks with the masks from the test set. The deforestation regions of interest for the problem occupy small areas compared to the rest, so we can consider the data to be imbalanced~\citep{pozzobon20}.

Each pixel of the test set mask is compared to its counterpart predicted by a model, defining the following measures: (i) true positives (TP): positive values that the system judged as positive (correct); (ii) false negatives (FN): positive values that the system judged as negative (error); (iii) true negatives (TN): negative values that the system judged as negative (correct); and (iv) false positives (FP): negative values that the system judged as positive (error).

With these values, we can calculate the following metrics:

\begin{itemize}
	
\item Accuracy: proportion of correct predictions, regardless of whether they are positive or negative. This measure is highly susceptible to dataset imbalances and can easily lead to a misleading conclusion about the system's performance.
\begin{equation*}
\text{Accuracy} = \small{\frac{\text{TP + TN}}{\text{TP + FP + TN + FN}}}\raisepunct{.}
\end{equation*}
	
\item Precision: the rate at which all samples classified as positive are actually positive. No negative sample is considered.
\begin{equation*}
\text{Precision} = \small{\frac{\text{TP}}{\text{TP + FP}}}\raisepunct{.}
\end{equation*}
	
\item Recall: the rate at which the system classifies as positive all samples that are truly positive. No positive sample is disregarded. Also known as sensitivity.
\begin{equation*}
\text{Recall} = \small{\frac{\text{TP}}{\text{TP + FN}}}\raisepunct{.}
\end{equation*}
	
\item F1-score: harmonic mean between precision and recall, providing a single value to indicate the model's overall performance.
\begin{equation*}
\text{F1} = \small{\frac{2 \text{TP}}{2 \text{TP + FP + FN}}}\raisepunct{.}
\end{equation*}
	
\item Cohen's Kappa coefficient: a measure of the agreement rate between two classifiers, prediction and annotation, beyond what is expected by chance.
\begin{equation*}
\kappa = \small{\frac{P_o - P_e}{1 - P_e}}\raisepunct{,}
\end{equation*}
\noindent where $P_o$ is the observed agreement between the evaluators, and $P_e$ is the agreement expected by chance. For the binary case, it can be defined as:
\begin{equation*}
\kappa = \small{\frac{2 \left( \text{TP} \cdot \text{TN} - \text{FN} \cdot \text{FP} \right)\raisepunct{,}}{\left( \text{TP} + \text{FP} \right) \left( \text{FP} + \text{TN} \right) \left( \text{TP} + \text{FN} \right) \left( \text{FN} + \text{TN} \right)}}\raisepunct{.}
\end{equation*}
	
\item Intersection over Union (IoU): overlap ratio between the regions correctly identified as positive by the system and those defined as positive by the reference mask or by the model itself.
\begin{equation*}
\text{IoU} = \small{\frac{\text{TP}}{\text{TP} + \text{FP} + \text{FN}}}\raisepunct{.}
\end{equation*}
	
\end{itemize}

\subsection{Model Adaptation}

As presented in Section~\ref{sec:related_work}, there are different ways to create a change detection model, we consider the Early Fusion strategy to be the most versatile. For this purpose, all segmentation models were adapted to receive an input of 16 bands, corresponding to the concatenation of the data from the two analyzed time instants. In our experiments, we initially used the UNet~\citep{unet} for proof-of-concept tests, but in later stages, we opted not to test it further due to computational cost, giving preference to more recent models such as UNet\texttt{++}. In addition, we explored other modern segmentation models, such as MultiResUNet, TransUNet (R50-ViT-B16), and SwinUNETR-V2.

For TransUNet (R50-ViT-B16), training a Vision Transformer (ViT) from scratch is computationally expensive, requiring large amounts of data and resources. Therefore, we chose to use pretrained weights from ImageNet~\citep{imagenet}, which allowed for a more efficient model fine-tuning with our reduced dataset. However, this approach has its limitations, since the domain of ImageNet images (natural photographs) is significantly different from the domain of the remote sensing images used in our work. This domain difference can impact the model's ability to generalize to our specific problem.

\subsection{Ensemble of Models}
\label{sec:ens_method}

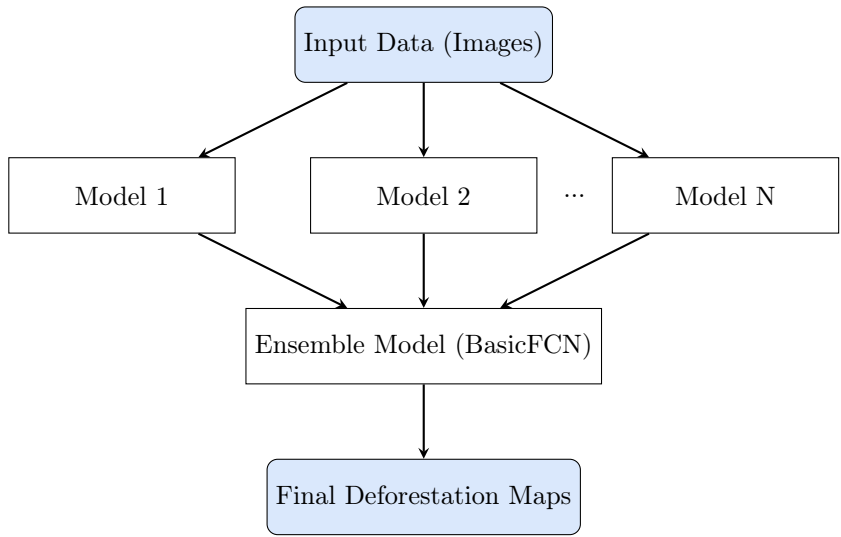
\begin{figure*}[!htb]
\centering
\begin{tikzpicture}[node distance=2cm, scale=0.7]
		
	\node (start) [startstop] {Input Data (Images)};
	\node (model1) [process, below of=start, xshift=-4cm] {Model 1};
	\node (model2) [process, below of=start, xshift=0cm] {Model 2};
	\node (dots) [below of=start, xshift=2cm] {...};
	\node (modeln) [process, below of=start, xshift=4cm] {Model N};
		
	\node (ensemble) [process, below of=model2, yshift=0] {Ensemble Model (BasicFCN)};
	\node (output) [startstop, below of=ensemble] {Final Deforestation Maps};
		
	\draw [arrow] (start) -- (model1);
	\draw [arrow] (start) -- (model2);
	\draw [arrow] (start) -- (modeln);
	
	\draw [arrow] (model1) -- (ensemble);
	\draw [arrow] (model2) -- (ensemble);
	\draw [arrow] (modeln) -- (ensemble);
		
	\draw [arrow] (ensemble) -- (output);
		
\end{tikzpicture}
\caption{Processing flow diagram with detection models and ensemble.}
\label{fig:ensemble_diagram}
\end{figure*}

As indicated in the diagram in Figure~\ref{fig:ensemble_diagram}, the predictions of multiple change detection models are combined through an ensemble model, with the objective of increasing overall effectiveness by weighting each one's decision. In this section, we will address how the combination of models is used in the context of the project and the nature of the model itself. We evaluate 3 ways to combine the models.

\subsubsection{Simple Voting}

The first approach, we will refer to as simple voting, follows a majority-vote logic among the models. Based on the results presented in Table~\ref{tab:magenta_results} and the binarization thresholds defined on the validation set, all models received the same sample from the dataset, generating binarized predictions according to the established threshold. The ensemble result is obtained through voting, where each pixel is classified as positive or negative according to the decision of the majority of models. In case of a tie, the negative result is chosen.

Mathematically, the decision for a pixel \( p \) can be expressed as:
\begin{equation}
\text{Class}(p) = \begin{cases}
1 & \text{if } \displaystyle \sum_{i=1}^{N} \text{Pred}_i(p) > \frac{N}{2}, \\
0 & \text{otherwise},
\end{cases}
\label{eq:votacao_simples}
\end{equation}
\noindent where $N$ is the number of models, $\text{Pred}_i(p)$ is the binarized prediction (0 or 1) of the $i$-th model for pixel $p$, and $\text{Class}(p)$ is the final class assigned to the pixel.

However, this strategy presents some limitations. One of them is that, when binarizing the predictions of each model, we lose valuable information about the probabilities assigned by the models for each outcome. Furthermore, the tie issue is problematic, especially when there are only four models, as it may occur more frequently. In such cases, it is not clear whether opting for the negative result instead of the positive one is the best decision.

\subsubsection{Probability-Weighted Voting}

With this in mind, a second ensemble approach emerges, called probability-weighted voting, which is slightly more refined. We still follow a voting idea, but without prior binarization of the results. In this approach, the non-binarized outputs of each model are averaged for each pixel of the deforestation map, and binarization is applied afterward to this average. To define the binarization threshold, we use the mean of the thresholds chosen for each model on the validation set.

Mathematically, the decision for a pixel $p$ is given by:
\begin{equation}
\text{Class}(p) = \begin{cases}
1 & \text{if } \displaystyle \frac{1}{N} \sum_{i=1}^{N} \text{Prob}_i(p) > \tau, \\
0 & \text{otherwise},
\end{cases}
\label{eq:votacao_ponderada}
\end{equation}
\noindent where $\text{Prob}_i(p)$ is the probability assigned by the $i$-th model for pixel $p$, and $\tau$ is the binarization threshold, calculated as the mean of the individual model thresholds:
\begin{equation}
\tau = \frac{1}{N} \sum_{i=1}^{N} \tau_i,
\label{eq:limiar_medio}
\end{equation}
\noindent and $\text{Class}(p)$ is the final class assigned to the pixel.

This way, we preserve the probabilistic information of the models and avoid problems associated with premature binarization. Although a tie is still possible, the chance is significantly lower, making the approach more robust and reliable.

However, this second approach also presents an important limitation: it assigns equal weights to all models, ignoring differences in their effectiveness. We know that some models are more effective than others, so it may not make sense to give the same relevance to each model in the final decision. A possible solution would be to weight both the probability average and the threshold average based on each model's effectiveness on the validation set. However, this would require defining a metric to serve as a weight, which in turn introduces new limitations. For example, if we choose the F1-score as the weighting metric, a model with high recall and low precision could receive a low weight, thus making little use of the model in the final decision, even though high recall could be beneficial for the ensemble. For these reasons, the third approach was proposed.

\subsubsection{Fully Convolutional Network}

Based on everything discussed previously, we decided that the model itself should determine the best way to combine the outputs of the previous models. For this, we chose to train a fully convolutional network with only two convolutional layers. We opted for a lightweight architecture because all the work of generating the deforestation maps is carried out by the individual models, and the ensemble network only needs to find a good way to combine them. The use of two convolutional layers allows the introduction of non-linearity, which helps improve the combination of predictions. Due to its nature, we call this model BasicFCN.

The architecture of BasicFCN is described in Table~\ref{tab:basic_fcn_architecture}, which details each layer and its parameters. The network consists of an initial convolutional layer, followed by batch normalization, a ReLU (Rectified Linear Unit) activation function, and a final convolutional layer to produce the output with a sigmoid function. In the results, we will include both the individual effectiveness of the models and the results obtained after combining all of them.

\begin{table*}[!htb]
\centering
\small
\setlength{\tabcolsep}{4mm} 
\caption{Architecture of BasicFCN. Each row describes a network layer and its parameters. Here, $N$ represents the number of models that make up the ensemble. All convolutions use stride 1, have no dilation, and include bias. The model preserves the height and width of the deforestation maps, reducing only the number of channels to 1.}
\label{tab:basic_fcn_architecture}
\begin{tabular}{l c c c c }
\toprule
\textbf{Layer}                  & \textbf{\makecell[l]{Input}} & \textbf{\makecell[l]{Output}} & \textbf{\makecell[l]{Kernel}} & \textbf{Pad} \\
\midrule
Convolution 1                   & \texttt{$N$}  & 12  & 3  & 1 \\
Batch Normalization              & 12           & 12  & -  & - \\
ReLU                             & -            & -   & -  & - \\
Convolution 2                    & 12           & 1   & 3  & 1 \\
\bottomrule
\end{tabular}
\end{table*}

\section{Experimental Results}
\label{cap:results}

This section presents the experiments conducted throughout this work, detailing the configurations evaluated in each case, the results obtained, and the relevant discussions. In addition, the limitations of the models and their comparison with other works from the literature are addressed. Both the individual detection models and the effectiveness of the ensemble approach are discussed.

\subsection{Baseline Results}
\label{sec:baseline_results}

This subsection presents the results of the simplest approach, which will serve as a baseline for model comparison. For each detection model, an initial evaluation was performed using default settings, without specific adjustments or advanced pre- or post-processing techniques. These results provide a starting point for comparative analyses and allow identifying the potential for improvements in each model.

\begin{table*}[!htb]
\centering
\setlength{\tabcolsep}{2.0mm}
\renewcommand{\arraystretch}{1}
\caption{Results of deforestation detection models using default settings. All metrics are presented as percentages (\%), and the threshold refers to the binarization threshold used to define positive and negative classes in the model outputs.}
\label{tab:baseline_results}
\small
\begin{tabular}{lccccccc}
\toprule
\textbf{Model} & \textbf{Threshold} & \textbf{Accuracy} & \textbf{IoU} & \textbf{Precision} & \textbf{Recall} & \textbf{F1-Score} & \textbf{Kappa} \\
\midrule
UNet\texttt{++} & 0.45 & 99.16 & 42.88 & 50.09 & 74.87 & 60.03 & 59.62 \\
MultiResUNet & 0.50 & 99.51 & 48.48 & \textbf{81.77} & 54.36 & 65.30 & 65.07 \\
SwinUNETR-V2 & 0.45 & 99.54 & 53.02 & 80.06 & 61.09 & 69.30 & 69.07 \\
TransUNet & 0.45 & \textbf{99.60} & \textbf{59.58} & 80.72 & \textbf{69.46} & \textbf{74.67} & \textbf{74.47} \\
\bottomrule
\end{tabular}
\end{table*}

At this stage, we used the training parameters described in Subsection~\ref{sec:train-model}, without employing additional pre- or post-processing techniques on the deforestation images or masks. The only exception was a simple normalization of the images, corresponding to the format in which they were stored, and the inclusion of the calculation of the additional NDVI channel, which is also part of the saved database. At the same time, the data augmentation techniques described in Subsection~\ref{sec:data-aug} were fully incorporated, as the variability they introduced proved essential — in some cases, their absence even prevented model convergence.

In light of these observations, we report the performance metrics of the models in Table~\ref{tab:baseline_results}. Note that the TransUNet model achieved the best overall performance, with emphasis on the F1-score (74.67\%) and IoU (59.58\%), indicating a better segmentation ability and accuracy in deforestation detection. The SwinUNETR-V2 also presented competitive results, with an IoU of 53.02\% and an F1-score of 69.30\%, suggesting a good balance between precision and recall. On the other hand, UNet\texttt{++} had the lowest precision (50.09\%) among the models, despite a relatively high recall (74.87\%), which may indicate a tendency towards false positives. Meanwhile, MultiResUNet showed high precision (81.77\%), but lower recall (54.36\%), suggesting possible under-detection of deforested areas. We can say that these results were marked by a strong imbalance between the precision and recall of the models.

\begin{figure}[!htb]
\centering
\subfloat[2017]{\includegraphics[width=0.32\linewidth]{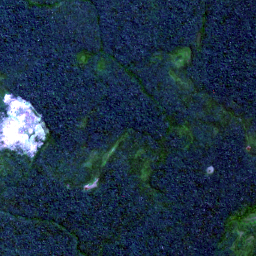}} \hfill
\subfloat[2018]{\includegraphics[width=0.32\linewidth]{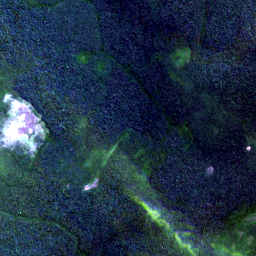}} \hfill
\subfloat[IoU]{\includegraphics[width=0.32\linewidth]{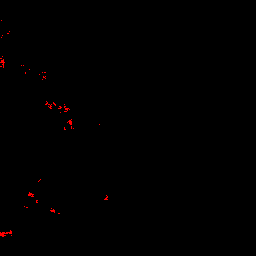}}
\caption{Comparison between two satellite images of the same location, taken one year apart (2017 and 2018), together with the IoU visualization for the SwinUNETR-V2 model. The red pixels represent false negatives of the model.}
\label{fig:small_area_problem}
\end{figure}

We performed a qualitative analysis of the results obtained at this stage, aiming to identify the behavior of the models' predictions and common patterns in the images associated with errors. Figure~\ref{fig:small_area_problem} illustrates one such case: we identified examples of false positives where the model incorrectly classified small regions with no evidence of forest degradation as new deforested areas. This type of pattern usually occurred, but not exclusively, when there was a significant disturbance in the image capture conditions. As in the cited figure, where the presence of a cloud is noticeable. This observation motivated the implementation of post-processing to remove small deforested regions, as described in Section~\ref{sec:mask_post_processing}.

\begin{figure}[!htb]
\centering
\subfloat[2017]{\includegraphics[width=0.32\linewidth]{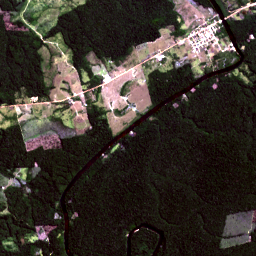}} \hfill
\subfloat[2018]{\includegraphics[width=0.32\linewidth]{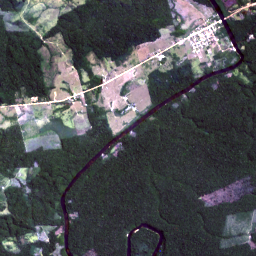}} \hfill
\subfloat[IoU]{\includegraphics[width=0.32\linewidth]{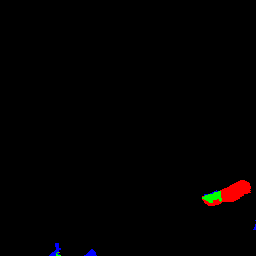}}
\caption{Comparison between two satellite images of the same location, taken one year apart (2017 and 2018), along with the IoU visualization for the SwinUNETR-V2 model. The green pixels represent the intersection; blue pixels, false negatives; and red pixels, false positives.}
\label{fig:mask_problem}
\end{figure}

Another issue that became evident with this preliminary analysis was the inconsistencies in the generated deforestation masks, as observed in Figure~\ref{fig:mask_problem}. It is noticeable that there was a large region of false negatives; however, upon analyzing the two collected images, the model's prediction appears to be more accurate than the ground truth mask. Indeed, this behavior was repeated on other occasions throughout the project. Among the possible explanations for this error, we list:

\begin{enumerate}
	
\item an error in the PRODES mask, which we consider unlikely, especially for the referenced image, since deforestation was identified, albeit in a smaller region;
	
\item an inconsistency in the image capture date relative to the date used by PRODES;
	
\item cases of progressive forest degradation, in which there is a change in the forest pattern but without complete canopy removal, which can lead to differing classifications between the models and PRODES experts.
	
\end{enumerate}

Finally, Table~\ref{tab:model_params} contains information about the total number of parameters and the memory size occupied by these models. At first glance, it appears that the larger number of parameters in the SwinUNETR-V2 and TransUNet models is directly related to their higher effectiveness, but this relationship can also be explained by the presence of pretraining in the case of TransUNet, which may have favored its results. In later sections, we will show that this greater number of parameters is not necessarily indicative of superior performance. In fact, by improving the overall pipeline, the results of other models, such as UNet\texttt{++} and MultiResUNet, can be significantly enhanced, achieving performance comparable to the more complex models, or even surpassing them in some scenarios.

\begin{table*}[!htb]
\centering
\setlength{\tabcolsep}{3.0mm}
\caption{Information on the total number of parameters and parameter size of the evaluated models.}
\label{tab:model_params}
\begin{tabular}{lcc}
\toprule
\textbf{Model} & \textbf{Total Parameters} & \textbf{Parameter Size (MB)} \\
\midrule
UNet\texttt{++} & 9,167,172 & 36.67 \\
MultiResUNet & 7,252,921 & 29.01 \\
SwinUNETR-V2 & 28,673,131 & 114.63 \\
TransUNet & 105,362,769 & 420.66 \\
\bottomrule
\end{tabular}
\end{table*}

\subsection{Removal of Small Deforested Regions}
\label{sec:small_reg_rm_results}

The results in Table~\ref{tab:small_region_results} show that the removal of small deforested regions (Section~\ref{sec:mask_post_processing}) impacts the models' metrics in a predictable manner. As expected, there is a reduction in recall for all models since this post-processing filters out areas activated by the model. For example, in the SwinUNETR-V2 model, recall decreased from 61.09\% to 59.93\%, while in TransUNet the reduction was from 69.46\% to 68.98\%. Conversely, precision increased, as seen in MultiResUNet, which went from 81.77\% to 86.64\%. This behavior indicates that removing small regions reduces false positives to a greater extent than false negatives. Overall, this balance resulted in an improvement in the F1-score for most models, except for TransUNet, whose value remained virtually unchanged (74.67\% without removal and 74.61\% with removal).

\begin{table*}[!htb]
\centering
\setlength{\tabcolsep}{2.0mm}
\caption{Results of change detection models with and without removal of small regions (50 pixels). All metrics are presented as percentages (\%), and the threshold refers to the binarization threshold used to define positive and negative classes in the model outputs.}
\label{tab:small_region_results}
\small
\begin{tabular}{lccccccc}
\toprule
\textbf{Model} & \textbf{Threshold} & \textbf{Accuracy} & \textbf{IoU} & \textbf{Precision} & \textbf{Recall} & \textbf{F1-Score} & \textbf{Kappa} \\
\midrule
\multicolumn{8}{c}{\textbf{With removal of small regions}} \\
\midrule
UNet\texttt{++} & 0.45 & 99.17 & 43.09 & 50.59 & 74.40 & 60.23 & 59.82 \\
MultiResUNet & 0.50 & 99.54 & 49.28 & \textbf{86.64} & 53.33 & 66.02 & 65.80 \\
SwinUNETR-V2 & 0.45 & 99.55 & 52.76 & 81.52 & 59.93 & 69.08 & 68.86 \\
TransUNet & 0.45 & \textbf{99.60} & \textbf{59.50} & 81.23 & \textbf{68.98} & \textbf{74.61} & \textbf{74.41} \\
\midrule
\multicolumn{8}{c}{\textbf{Without removal of small regions}} \\
\midrule
UNet\texttt{++} & 0.45 & 99.16 & 42.88 & 50.09 & 74.87 & 60.03 & 59.62 \\
MultiResUNet & 0.50 & 99.51 & 48.48 & \textbf{81.77} & 54.36 & 65.30 & 65.07 \\
SwinUNETR-V2 & 0.45 & 99.54 & 53.02 & 80.06 & 61.09 & 69.30 & 69.07 \\
TransUNet & 0.45 & \textbf{99.60} & \textbf{59.58} & 80.72 & \textbf{69.46} & \textbf{74.67} & \textbf{74.47} \\
\bottomrule
\end{tabular}
\end{table*}

\subsection{Histogram Equalization}
\label{sec:hist_eq_results}

\begin{figure}[!htb]
\centering
\subfloat[Without equalization]{\includegraphics[width=0.35\linewidth]{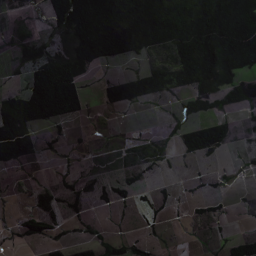}} \hspace{1em}
\subfloat[With equalization]{\includegraphics[width=0.35\linewidth]{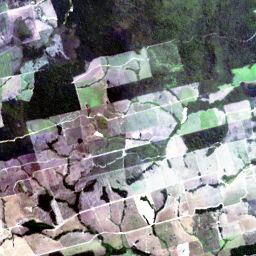}} \\
\subfloat[Without equalization]{\includegraphics[width=0.35\linewidth]{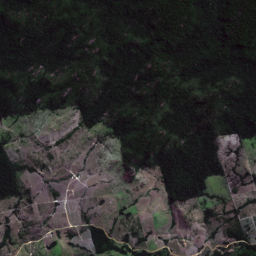}} \hspace{1em}
\subfloat[With equalization]{\includegraphics[width=0.35\linewidth]{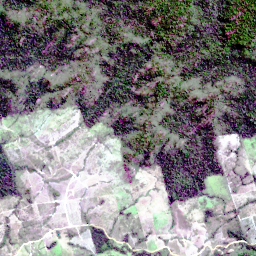}}
\caption{Comparison between satellite images without and with histogram equalization, visualized in the RGB channels.}
\label{fig:eq_hist_effect}
\end{figure}

This subsection addresses the effect that histogram equalization of the satellite images, as described in Subsection~\ref{sec:method-eq-hist}, has on the models' effectiveness. It is known from~\citep{prodes_method} that images undergo contrast enhancement before being photo-interpreted by PRODES experts. We then decided to evaluate the effect of histogram equalization on the images, as illustrated in Figure~\ref{fig:eq_hist_effect}.

The analysis of Figure~\ref{fig:eq_hist_effect} shows that histogram equalization resulted in a very strong contrast enhancement in the images, facilitating the identification of boundaries between preserved forest and deforested areas. In image (d), in particular, it is possible to notice forest degradation that was not evident in image (c), without equalization. However, the enhancement also amplified the presence of noise in the images, which may introduce additional challenges for analysis. It is uncertain whether lack of contrast would indeed be a problem for computer vision models, but it is possible that equalization has the effect of standardizing the values provided to the networks, improving the overall result.

To evaluate the impact of equalization on model performance, we conducted tests on all trained models. The test setup was the same as described in Section~\ref{sec:small_reg_rm_results}, with the only difference that the input images had their histograms equalized. The results are reported in Table~\ref{tab:eq_hist_results}, which presents the obtained metrics and a comparison with the previous test results.

\begin{table*}[!htb]
\centering
\setlength{\tabcolsep}{2.0mm}
\caption{Results of change detection models with and without histogram equalization. All metrics are presented as percentages (\%), and the threshold refers to the binarization threshold used to define positive and negative classes in the model outputs.}
\label{tab:eq_hist_results}
\small
\begin{tabular}{lccccccc}
\toprule
\textbf{Model} & \textbf{Threshold} & \textbf{Accuracy} & \textbf{IoU} & \textbf{Precision} & \textbf{Recall} & \textbf{F1-Score} & \textbf{Kappa} \\
\midrule
\multicolumn{8}{c}{\textbf{With histogram equalization}} \\
\midrule
UNet\texttt{++} & 0.45 & 99.59 & 60.69 & 76.45 & 74.64 & 75.53 & 75.33 \\
MultiResUNet & 0.50 & 99.53 & 57.88 & 70.99 & \textbf{75.80} & 73.32 & 73.08 \\
SwinUNETR-V2 & 0.40 & 99.59 & 59.64 & 77.27 & 72.33 & 74.72 & 74.51 \\
TransUNet & 0.40 & \textbf{99.60} & \textbf{61.15} & \textbf{78.19} & 73.73 & \textbf{75.89} & \textbf{75.70} \\
\midrule
\multicolumn{8}{c}{\textbf{Without histogram equalization}} \\
\midrule
UNet\texttt{++} & 0.45 & 99.17 & 43.09 & 50.59 & 74.40 & 60.23 & 59.82 \\
MultiResUNet & 0.50 & 99.54 & 49.28 & \textbf{86.64} & 53.33 & 66.02 & 65.80 \\
SwinUNETR-V2 & 0.45 & 99.55 & 52.76 & 81.52 & 59.93 & 69.08 & 68.86 \\
TransUNet & 0.45 & \textbf{99.60} & \textbf{59.50} & 81.23 & \textbf{68.98} & \textbf{74.61} & \textbf{74.41} \\
\bottomrule
\end{tabular}
\end{table*}

\subsection{Replacement of Clear-Cut Deforestation Patterns with Fire Use}
\label{sec:magenta_results}

This subsection addresses the issue of magenta regions observed in satellite images, which are likely associated with clear-cut deforestation patterns involving fire use. As discussed in Subsection~\ref{sec:magenta_method}, these regions represent a challenge for analysis, as they may be mistakenly interpreted as unchanged when in fact they correspond to newly deforested areas. To tackle this problem, we propose an approach to replace these areas with more common deforestation texture patterns, as detailed in the methodology.

\begin{figure}[!htb]
\centering
\subfloat[2018]{\includegraphics[width=0.3\linewidth]{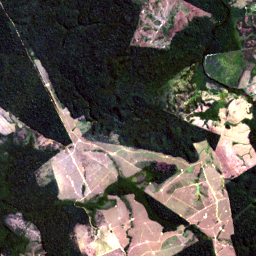}} \hfill
\subfloat[2019]{\includegraphics[width=0.3\linewidth]{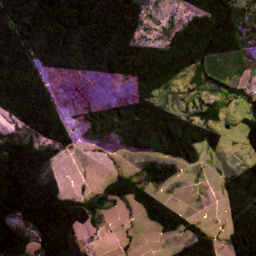}} \hfill
\subfloat[IoU]{\includegraphics[width=0.3\linewidth]{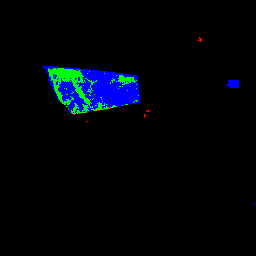}} \\
\subfloat[2018]{\includegraphics[width=0.3\linewidth]{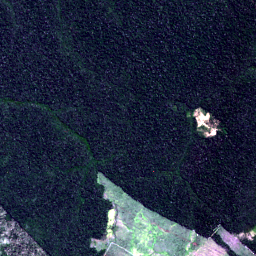}} \hfill
\subfloat[2019]{\includegraphics[width=0.3\linewidth]{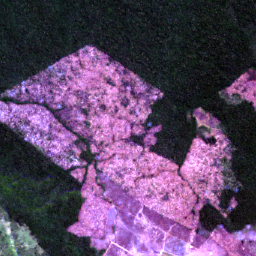}} \hfill
\subfloat[IoU]{\includegraphics[width=0.3\linewidth]{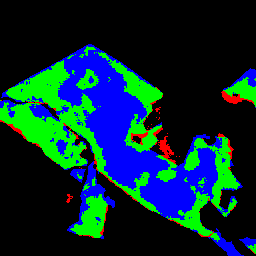}}
\caption{Examples of magenta regions in satellite images, their predictions, and the corresponding IoU maps. Green pixels represent the intersection; blue pixels, false negatives; and red pixels, false positives. The prediction of the SwinUNETR-V2 model from Section~\ref{sec:baseline_results} illustrates the difficulty in handling the magenta pattern.}
\label{fig:magenta_problem}
\end{figure}

As previously mentioned, the models struggle to handle the magenta pattern, producing false negatives. This problem is illustrated in Figure~\ref{fig:magenta_problem}, where the prediction of the SwinUNETR-V2 model from Section~\ref{sec:baseline_results} fails to correctly identify the affected regions. The figure shows examples of satellite images with magenta areas, their predictions, and the corresponding IoU maps, highlighting false negatives in blue.

To evaluate the impact of replacing the magenta regions, we performed a test identical to that described in Section~\ref{sec:hist_eq_results}, with the only difference being that the input images were preprocessed to replace clear-cut deforestation areas affected by fire use with a more common deforestation texture, as detailed in Subsection~\ref{sec:magenta_method}. The results of this test are reported in Table~\ref{tab:magenta_results}, which compares the obtained metrics with those from Section~\ref{sec:hist_eq_results}, allowing a direct analysis of the effect of this approach.

\begin{table*}[!htb]
\centering
\setlength{\tabcolsep}{2.0mm}
\caption{Results of change detection models with and without replacement of magenta regions. All metrics are presented as percentages (\%), and the threshold refers to the binarization threshold used to define positive and negative classes in the model outputs.}
\label{tab:magenta_results}
\small
\begin{tabular}{lccccccc}
\toprule
\textbf{Model} & \textbf{Threshold} & \textbf{Accuracy} & \textbf{IoU} & \textbf{Precision} & \textbf{Recall} & \textbf{F1-Score} & \textbf{Kappa} \\
\midrule
\multicolumn{8}{c}{\textbf{With magenta region replacement}} \\
\midrule
UNet\texttt{++} & 0.50 & 99.62 & 63.37 & 78.08 & 77.08 & 77.58 & 77.39 \\
MultiResUNet & 0.45 & 99.57 & 59.78 & 73.21 & 76.52 & 74.83 & 74.61 \\
SwinUNETR-V2 & 0.40 & \textbf{99.63} & \textbf{64.20} & \textbf{77.25} & \textbf{79.17} & \textbf{78.20} & \textbf{78.01} \\
TransUNet & 0.45 & 99.61 & 62.44 & 77.07 & 76.69 & 76.88 & 76.68 \\
\midrule
\multicolumn{8}{c}{\textbf{Without magenta region replacement}} \\
\midrule
UNet\texttt{++} & 0.45 & 99.59 & 60.69 & 76.45 & 74.64 & 75.53 & 75.33 \\
MultiResUNet & 0.50 & 99.53 & 57.88 & 70.99 & \textbf{75.80} & 73.32 & 73.08 \\
SwinUNETR-V2 & 0.40 & 99.59 & 59.64 & 77.27 & 72.33 & 74.72 & 74.51 \\
TransUNet & 0.40 & \textbf{99.60} & \textbf{61.15} & \textbf{78.19} & 73.73 & \textbf{75.89} & \textbf{75.70} \\
\bottomrule
\end{tabular}
\end{table*}

The results presented in Table~\ref{tab:magenta_results} show that replacing magenta regions led to a generalized improvement in metrics for all evaluated models, except for the precision of TransUNet, which showed a slight decrease. This improvement is particularly evident in the F1-score metric, where, for example, SwinUNETR-V2 improved from 74.72\% to 78.20\%, and TransUNet from 75.89\% to 76.88\%. Furthermore, a notable outcome of this test was the superior performance of SwinUNETR-V2 compared to TransUNet, which had not been observed in previous experiments. This fact makes the model especially interesting since, besides not being pretrained, it has fewer parameters than TransUNet, as illustrated in Table~\ref{tab:model_params}.

\subsection{Ensemble of Models}

In this subsection, we discuss three strategies of combining the previously addressed models, all of them  described in Section~\ref{sec:ens_method}, and their results. We expect that combining the models through these ensemble methods will result in performance superior to each individual model, leveraging the strengths of each approach while minimizing their limitations.

\begin{table*}[!htb]
\centering
\setlength{\tabcolsep}{2.0mm}
\caption{Results of ensemble approaches and base models. All metrics are presented as percentages (\%), and the threshold refers to the binarization threshold used to define the positive and negative classes in the model outputs.}
\label{tab:ens_results}
\small
\begin{tabular}{lccccccc}
\toprule
\textbf{Model} & \textbf{Threshold} & \textbf{Accuracy} & \textbf{IoU} & \textbf{Precision} & \textbf{Recall} & \textbf{F1-Score} & \textbf{Kappa} \\
\midrule
\multicolumn{8}{c}{\textbf{Ensemble Approaches}} \\
\midrule
Simple Voting & - & 99.66 & 65.51 & \textbf{83.32} & 75.40 & 79.16 & 78.99 \\
Weighted Voting & - & 99.66 & 66.07 & 79.57 & \textbf{79.57} & 79.57 & 79.39 \\
\textit{BasicFCN} & 0.45 & \textbf{99.68} & \textbf{67.24} & 83.13 & 77.87 & \textbf{80.41} & \textbf{80.25} \\
\midrule
\multicolumn{8}{c}{\textbf{Base Models}} \\
\midrule
UNet\texttt{++} & 0.50 & 99.62 & 63.37 & 78.08 & 77.08 & 77.58 & 77.39 \\
MultiResUNet & 0.45 & 99.57 & 59.78 & 73.21 & 76.52 & 74.83 & 74.61 \\
SwinUNETR-V2 & 0.40 & \textbf{99.63} & \textbf{64.20} & \textbf{77.25} & \textbf{79.17} & \textbf{78.20} & \textbf{78.01} \\
TransUNet & 0.45 & 99.61 & 62.44 & 77.07 & 76.69 & 76.88 & 76.68 \\
\bottomrule
\end{tabular}
\end{table*}

Overall, the ensemble approaches outperformed the individual base models (Table~\ref{tab:ens_results}), although not consistently across all metrics. The Fully Convolutional Network (FCN) achieved the highest accuracy (99.68\%), F1-score (80.41\%), and IoU (67.24\%), surpassing all base models. Weighted voting attained the highest recall (79.57\%) among the ensemble methods, while simple voting stood out for the highest precision (83.32\%). Among the base models, SwinUNETR-V2 came closest to the ensemble performance, with an F1-score of 78.20\% and IoU of 64.20\%.

\subsection{Discussion}

In this subsection, we discuss the results obtained in the previous ones, compare them with the literature, and consider the relationships between effectiveness and efficiency of the evaluated models and ensemble approaches. The analysis covers both the performance metrics and the associated computational cost, aiming for a balance between accuracy and practical feasibility.

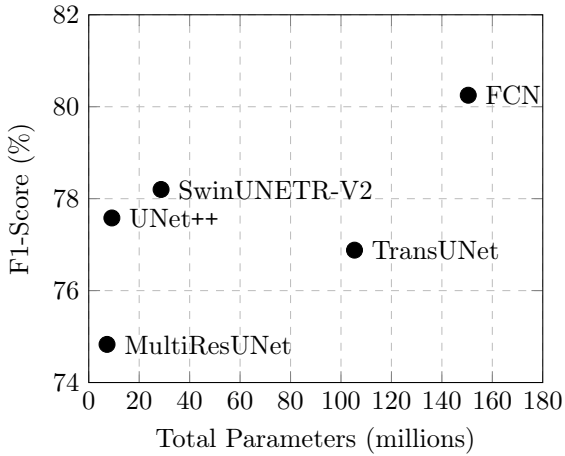
\begin{figure}[!htb]
\centering
\begin{tikzpicture}
	\begin{axis}[
		xlabel={Total Parameters (millions)},
		ylabel={F1-Score (\%)},
		xmin=0, xmax=180,
		ymin=74, ymax=82,
		xtick={0, 20, 40, 60, 80, 100, 120, 140, 160, 180},
		ytick={74, 76, 78, 80, 82},
		legend pos=north east,
		ymajorgrids=true,
		xmajorgrids=true,
		grid style=dashed,
		width=\linewidth,
		height=0.85\linewidth,
		nodes near coords,
		point meta=explicit symbolic,
		every node near coord/.append style={anchor=west, xshift=3pt} 
		]
			
		\addplot[
		only marks,
		mark=*,
		mark size=3pt,
		] coordinates {
			(7.252921, 74.83) [MultiResUNet]
			(9.167172, 77.58) [UNet\texttt{++}]
			(28.673131, 78.20) [SwinUNETR-V2]
			(105.362769, 76.88) [TransUNet]
			(150.456561, 80.25) [FCN]
		};
		
	\end{axis}
\end{tikzpicture}
\caption{Relationship between total model parameters and F1-score. Points represent individual models and the FCN ensemble approach, with model names positioned to the right of each point. For the FCN, the parameter count corresponds to the sum of all other models plus the network combining the results.}
\label{fig:f1_vs_params}
\end{figure}

The plot in Figure~\ref{fig:f1_vs_params} shows the relationship between model parameter counts and obtained F1-scores, allowing a comparison between effectiveness and computational cost. It is observed that, in general, models with more parameters tend to achieve better performance, as seen in SwinUNETR-V2 and the FCN ensemble. However, TransUNet, which has the highest parameter count among individual models, shows a lower F1-score than these models. In previous tests, TransUNet was usually the best in F1-score, but as we refined the pipeline, the other models benefited more and surpassed it. This behavior may be a consequence of TransUNet's higher complexity, which makes it harder to train properly; the fact that it is the only model used in its pretrained version supports this hypothesis, since pretraining explains its superior effectiveness in simpler methodologies.

To contextualize our results, we will compare the effectiveness achieved by models from the literature. However, it is important to emphasize that, although the models from the literature serve as references for their respective effectiveness, a direct comparison between them is unfeasible due to the lack of standardization in the test sets used. Each study employs different datasets and evaluation methodologies, which hinders the generalization of results. In contrast, our tests were conducted on the same dataset, ensuring consistency in the evaluation. We combine our results with those from the literature models in Table~\ref{tab:literature_results}, allowing a comparative analysis within these limitations.

The reproduction of results from the literature is also difficult to achieve, since most works do not provide code or pre-trained models. Moreover, the papers usually omit important details about the dataset generation process. The work of~\citet{pozzobon20} reported very high effectiveness in the problem, even when compared with more recent works in the literature. A possible explanation for this difference lies in the dataset construction, where manual adjustments applied to the data may have contributed to superior results. However, this cannot be verified at the moment without access to the dataset itself.

Another model that stood out was ChangeFormer~\citep{amazon_changeformer}, which achieved very high metrics, reaching an F1-score of 84.62\%. This work employs preprocessing strategies and dataset construction methods that could be integrated into the current approach to potentially improve results, as discussed in Section~\ref{sec:related_work}. Nevertheless, once again, direct comparison between models is hampered by differences in dataset construction. In particular, the authors filtered the dataset to reduce class imbalance, removing samples containing only a single class and discarding those in which the ``change'' class represented less than 10\% of the total image area. This strategy mitigates extreme imbalances between change and no-change classes but also introduces bias into the dataset, which is not present in the dataset used in this work. Therefore, the observed differences in results may be partially attributed to this discrepancy in data distribution, reinforcing the need to consider the dataset context when interpreting performance metrics.

\begin{table*}[!htb]
\centering
\setlength{\tabcolsep}{3.0mm}
\caption{Comparison of different models from the literature with the effectiveness obtained in the tests. Bold names represent models from the current work. All values are presented as percentages.}
\label{tab:literature_results}
\small
\begin{tabular}{llcccc}
\toprule
\textbf{Model}                                      & \textbf{Year} & \textbf{IoU} & \textbf{Precision} & \textbf{Recall} & \textbf{F1-Score} \\
\midrule
ResUNet~\citep{pozzobon20}                          & 2020         & 94.87        & 93.58             & 95.74           & 94.65             \\
EF~\citep{cd_deeplab}                               & 2022         & -            & -                 & -               & 63.24             \\
S-CNN~\citep{cd_deeplab}                            & 2022         & -            & -                 & -               & 62.94             \\
DLCD-14~\citep{cd_deeplab}                          & 2022         & -            & 74.98             & 72.15           & 73.42             \\
LSTM\texttt{+}UNet (BA)~\citep{cd_cerrado}          & 2022         & -            & -                 & -               & 69.54             \\
LSTM\texttt{+}UNet (MT)~\citep{cd_cerrado}          & 2022         & -            & -                 & -               & 71.28             \\
U-TAE (8 \textit{head})~\citep{utae}                & 2023         & 47.3         & 70.9              & 58.6            & 64.2              \\
ChangeFormer~\citep{amazon_changeformer}            & 2024         & 73.33        & 84.70             & 84.53           & 84.62             \\
\textbf{MultiResUNet}                               & 2025         & 59.78        & 73.21             & 76.52           & 74.83             \\
\textbf{TransUNet}                                  & 2025         & 62.44        & 77.07             & 76.69           & 76.88             \\
\textbf{UNet\texttt{++}}                            & 2025         & 63.37        & 78.08             & 77.08           & 77.58             \\
\textbf{SwinUNETR-V2}                               & 2025         & 64.20        & 77.25             & 79.17           & 78.20             \\
\textbf{FCN}                                        & 2025         & 67.24        & 83.13             & 77.87           & 80.41             \\
\bottomrule
\end{tabular}
\end{table*}

\section{Conclusions}

In this work, we addressed the problem of deforestation detection and segmentation using change detection techniques. For this purpose, we used Landsat 8 satellite images combined with deforestation masks from PRODES, which provide precise temporal information about land cover. The construction of this dataset was fundamental for the analysis and evaluation of the proposed models. The evaluated models are segmentation neural networks applied to the change detection task. We included both classical and modern networks, ranging from fully convolutional models to networks incorporating self-attention mechanisms from Transformer models, enabling a comprehensive assessment of different approaches.

Direct comparison between our models and those from the literature is quite limited, as there is no standardization in the datasets used, as previously discussed. Nevertheless, the obtained effectiveness measures indicate that we achieved competitive values relative to the state of the art. As shown in Table~\ref{tab:literature_results}, our models obtained promising results, with emphasis on the ensemble approach, which reached an F1-score of 80.41\%. However, there is still room for improvement, especially compared to more recent models that have achieved F1-scores above 84\%. It is important to highlight, however, that these values were obtained on filtered datasets to avoid extreme imbalance, which was not performed in our tests as it could add an undesired bias to our methods, possibly explaining part of the observed difference.

One of the main directions for future work would be to evaluate whether the proposed models could be applied to other Brazilian biomes. The Cerrado, for example, is a natural candidate for such investigation. Besides being one of the most threatened biomes by human activity~\citep{cerrado_policy}, it occupies a vast area of the national territory and shelters thousands of endemic species~\citep{cd_cerrado}, making its preservation of utmost importance. Another topic worthy of investigation is the use of other loss functions for model optimization. Also, another less explored line is how to deal with cloud occlusion, which represents one of the main challenges for deforestation detection.


\begin{thebibliography}{49}
\providecommand{\natexlab}[1]{#1}
\providecommand{\url}[1]{{#1}}
\providecommand{\urlprefix}{URL }
\providecommand{\doi}[1]{\url{https://doi.org/#1}}
\providecommand{\eprint}[2][]{\url{#2}}
 \bibcommenthead

\bibitem[{sha(1998)}]{shapefile}
 (1998) {ESRI Shapefile Technical Description}. Environmental Systems Research
  Institute

\bibitem[{lan(2019)}]{landsat_manual}
 (2019) {Landsat 9 (L8) Data User Handbook}. United States Geological Survey,
  \urlprefix\url{https://www.usgs.gov/media/files/landsat-8-data-users-handbook},
  version 5.0

\bibitem[{pro(2022)}]{prodes_method}
 (2022) {Metodologia Utilizada nos Sistemas PRODES e DETER}. Instituto Nacional
  de Pesquisas Espaciais, \url{http://urlib.net/8JMKD3MGP3W34T/47GAF6S}

\bibitem[{Abraham and Khan(2019)}]{focal_tversky}
Abraham N, Khan NM (2019) {A Novel Focal Tversky Loss Function With Improved
  Attention U-Net for Lesion Segmentation}. In: IEEE 16th International
  Symposium on Biomedical Imaging, pp 683--687

\bibitem[{Alshehri et~al(2024)Alshehri, Ouadou, and
  Scott}]{amazon_changeformer}
Alshehri M, Ouadou A, Scott GJ (2024) {Deep Transformer-Based Network
  Deforestation Detection in the Brazilian Amazon Using Sentinel-2 Imagery}.
  IEEE Geoscience and Remote Sensing Letters 21:1--5

\bibitem[{Andrade et~al(2020)Andrade, Costa, Mota, Ortega, Feitosa, Soto, and
  Heipke}]{cd_amazon_comparison}
Andrade RB, Costa GAOP, Mota GLA, et~al (2020) {Evaluation of Semantic
  Segmentation Methods for Deforestation Detection in the Amazon}. The
  International Archives of the Photogrammetry, Remote Sensing and Spatial
  Information Sciences XLIII-B3-2020:1497--1505

\bibitem[{Andrade et~al(2022)Andrade, Mota, and Costa}]{cd_deeplab}
Andrade RB, Mota GLA, Costa GAOP (2022) {Deforestation Detection in the Amazon
  Using DeepLabv3\texttt{+} Semantic Segmentation Model Variants}. Remote
  Sensing 14(19)

\bibitem[{Bandara and Patel(2022)}]{change_former}
Bandara WGC, Patel VM (2022) {A Transformer-Based Siamese Network for Change
  Detection}. In: IEEE International Geoscience and Remote Sensing Symposium,
  pp 207--210

\bibitem[{de~Bem et~al(2020)de~Bem, de~Carvalho~Junior, Fontes~Guimarães, and
  Trancoso~Gomes}]{pozzobon20}
de~Bem PP, de~Carvalho~Junior OA, Fontes~Guimarães R, et~al (2020) {Change
  Detection of Deforestation in the Brazilian Amazon Using Landsat Data and
  Convolutional Neural Networks}. Remote Sensing 12(6)

\bibitem[{Boucher et~al(2013)Boucher, Roquemore, and
  Fitzhugh}]{deforestation_success}
Boucher D, Roquemore S, Fitzhugh E (2013) {Brazil's Success in Reducing
  Deforestation}. Tropical Conservation Science 6(3):426--445

\bibitem[{Chen et~al(2022)Chen, Qi, and Shi}]{rs_transformer}
Chen H, Qi Z, Shi Z (2022) {Remote Sensing Image Change Detection With
  Transformers}. IEEE Transactions on Geoscience and Remote Sensing 60:1--14

\bibitem[{Chen et~al(2021)Chen, Lu, Yu, Luo, Adeli, Wang, Lu, Yuille, and
  Zhou}]{transunet_ori}
Chen J, Lu Y, Yu Q, et~al (2021) {TransUNet: Transformers Make Strong Encoders
  for Medical Umage Segmentation}. arXiv preprint arXiv:210204306

\bibitem[{Chen et~al(2024)Chen, Mei, Li, Lu, Yu, Wei, Luo, Xie, Adeli, Wang,
  Lungren, Zhang, Xing, Lu, Yuille, and Zhou}]{transunet_new}
Chen J, Mei J, Li X, et~al (2024) {TransUNet: Rethinking the U-Net Architecture
  Design for Medical Image Segmentation through the Lens of Transformers}.
  Medical Image Analysis 97:103280

\bibitem[{Chen et~al(2025)Chen, Samat, {Maghsoodi Tilaki}, and
  Duan}]{luc_impact}
Chen M, Samat N, {Maghsoodi Tilaki} MJ, et~al (2025) {Land Use/Cover Change
  Simulation Research: A System Literature Review based on Bibliometric
  Analyses}. Ecological Indicators 170:112991

\bibitem[{Coulibaly(2025)}]{coulibaly2025effects}
Coulibaly Y (2025) {The Effects of Resource-Backed Loans on Deforestation:
  Evidence from Developing Countries}. World Development 188:106905

\bibitem[{Deng et~al(2009)Deng, Dong, Socher, Li, Li, and Fei-Fei}]{imagenet}
Deng J, Dong W, Socher R, et~al (2009) {ImageNet: A Large-Acale Hierarchical
  Image Database}. In: IEEE Conference on Computer Vision and Pattern
  Recognition, pp 248--255

\bibitem[{Ellwanger et~al(2020)Ellwanger, Kulmann-Leal, Kaminski,
  Valverde-Villegas, Veiga, Spilki, Fearnside, Caesar, Giatti, and
  Wallau}]{amazon_disease}
Ellwanger JH, Kulmann-Leal B, Kaminski VL, et~al (2020) {Beyond Diversity Loss
  and Climate Change: Impacts of Amazon Deforestation on Infectious Diseases
  and Public Health}. Anais da Academia Brasileira de Ci{\^e}ncias 92

\bibitem[{Fisher et~al(2005)Fisher, Comber, and Wadsworth}]{land_use_cover}
Fisher P, Comber AJ, Wadsworth R (2005) {Land Use and Land Cover: Contradiction
  or Complement}. Re-presenting GIS 85:98

\bibitem[{He et~al(2016)He, Zhang, Ren, and Sun}]{resnet}
He K, Zhang X, Ren S, et~al (2016) Deep residual learning for image
  recognition. In: IEEE Conference on Computer Vision and Pattern Recognition

\bibitem[{He et~al(2023)He, Nath, Yang, Tang, Myronenko, and Xu}]{swinunetrv2}
He Y, Nath V, Yang D, et~al (2023) {SwinUNETR-V2: Stronger Swin Transformers
  with Stagewise Convolutions for 3D Medical Image Segmentation}. In:
  Greenspan H, Madabhushi A, Mousavi P, et~al (eds) Medical Image Computing and
  Computer Assisted Intervention. Springer Nature Switzerland, Cham, pp
  416--426

\bibitem[{Heck et~al(2005)Heck, Loebens, and Carvalho}]{amazon_population}
Heck E, Loebens F, Carvalho PD (2005) {Amaz{\^o}nia Ind{\'\i}gena: Conquistas e
  Desafios}. Estudos Avan{\c{c}}ados 19:237--255

\bibitem[{Huang et~al(2021)Huang, Tang, Hupy, Wang, and Shao}]{ndvi}
Huang S, Tang L, Hupy JP, et~al (2021) {A Commentary Review on the Use of
  Normalized Difference Vegetation Index (NDVI) in the Era of Popular Remote
  Sensing}. Journal of Forestry Research 32(1):1--6

\bibitem[{Ibtehaz and Rahman(2020)}]{multiresunet}
Ibtehaz N, Rahman MS (2020) {MultiResUNet : Rethinking the U-Net Architecture
  for Multimodal Biomedical Image Segmentation}. Neural Networks 121:74--87

\bibitem[{{Instituto Nacional de Pesquisas Espaciais}(1999)}]{prodes}
{Instituto Nacional de Pesquisas Espaciais} (1999) {Programa de Monitoramento
  da Amazônia e Demais Biomas}.
  \url{http://terrabrasilis.dpi.inpe.br/downloads/}

\bibitem[{Karaman et~al(2023)Karaman, Sainte Fare~Garnot, and Wegner}]{utae}
Karaman K, Sainte Fare~Garnot V, Wegner JD (2023) {Deforestation Detection in
  the Amazon with Sentinel-1 SAR Image Time Series}. ISPRS Annals of the
  Photogrammetry, Remote Sensing and Spatial Information Sciences
  X-1/W1-2023:835--842

\bibitem[{Khelifi and Mignotte(2020)}]{cd_review_2022}
Khelifi L, Mignotte M (2020) {Deep Learning for Change Detection in Remote
  Sensing Images: Comprehensive Review and Meta-Analysis}. IEEE Access
  8:126385--126400

\bibitem[{Khorram et~al(2012)Khorram, Koch, Van~der Wiele, and
  Nelson}]{remote_sensing_book}
Khorram S, Koch FH, Van~der Wiele CF, et~al (2012) {Remote Sensing}. Springer
  Science \& Business Media

\bibitem[{Kingma(2014)}]{adam}
Kingma DP (2014) {Adam: A Method for Stochastic Optimization}. arXiv preprint
  arXiv:14126980 pp 1--15

\bibitem[{Lin et~al(2020)Lin, Goyal, Girshick, He, and Dollár}]{focal_loss}
Lin TY, Goyal P, Girshick R, et~al (2020) {Focal Loss for Dense Object
  Detection}. IEEE Transactions on Pattern Analysis and Machine Intelligence
  42(2):318--327

\bibitem[{Liu et~al(2021)Liu, Lin, Cao, Hu, Wei, Zhang, Lin, and
  Guo}]{swin_transformer}
Liu Z, Lin Y, Cao Y, et~al (2021) {Swin Transformer: Hierarchical Vision
  Transformer Using Shifted Windows}. In: IEEE/CVF International Conference on
  Computer Vision, pp 10012--10022

\bibitem[{Luiz and Steinke(2022)}]{cerrado_policy}
Luiz CHP, Steinke VA (2022) {Recent Environmental Legislation in Brazil and the
  Impact on Cerrado Deforestation Rates}. Sustainability 14(13)

\bibitem[{Luo et~al(2024)Luo, Quaas, and Han}]{luo2024decreased}
Luo H, Quaas J, Han Y (2024) {Decreased Cloud Cover Partially Offsets the
  Cooling Effects of Surface Albedo Change due to Deforestation}. Nature
  Communications 15(1):7345

\bibitem[{Maretto et~al(2021)Maretto, Fonseca, Jacobs, Körting, Bendini, and
  Parente}]{maretto21}
Maretto RV, Fonseca LMG, Jacobs N, et~al (2021) {Spatio-Temporal Deep Learning
  Approach to Map Deforestation in Amazon Rainforest}. IEEE Geoscience and
  Remote Sensing Letters 18(5):771--775

\bibitem[{Matosak et~al(2022)Matosak, Fonseca, Taquary, Maretto, Bendini, and
  Adami}]{cd_cerrado}
Matosak BM, Fonseca LMG, Taquary EC, et~al (2022) {Mapping Deforestation in
  Cerrado Based on Hybrid Deep Learning Architecture and Medium Spatial
  Resolution Satellite Time Series}. Remote Sensing 14(1)

\bibitem[{Patel and Goswami(2014)}]{hist_eq}
Patel S, Goswami M (2014) {Comparative Analysis of Histogram Equalization
  Techniques}. In: International Conference on Contemporary Computing and
  Informatics, pp 167--168

\bibitem[{Qin et~al(2025)Qin, Wang, Ziegler, Fu, and Zeng}]{qin2025impact}
Qin Y, Wang D, Ziegler AD, et~al (2025) {Impact of Amazonian Deforestation on
  Precipitation Reverses Between Seasons}. Nature 639(8053):102--108

\bibitem[{Ridnik et~al(2021)Ridnik, Ben-Baruch, Zamir, Noy, Friedman, Protter,
  and Zelnik-Manor}]{asymmetrical_focal_loss}
Ridnik T, Ben-Baruch E, Zamir N, et~al (2021) {Asymmetric Loss for Multi-Label
  Classification}. In: IEEE/CVF International Conference on Computer Vision, pp
  82--91

\bibitem[{Ronneberger et~al(2015)Ronneberger, Fischer, and Brox}]{unet}
Ronneberger O, Fischer P, Brox T (2015) {U-Net: Convolutional Networks for
  Biomedical Image Segmentation}. In: Navab N, Hornegger J, Wells WM, et~al
  (eds) Medical Image Computing and Computer-Assisted Intervention. Springer
  International Publishing, Cham, pp 234--241

\bibitem[{Schowengerdt(2006)}]{remote_sensing_models}
Schowengerdt RA (2006) {Remote Sensing: Models and Methods for Image
  Processing}. Elsevier

\bibitem[{Shi et~al(2020)Shi, Zhang, Zhang, Chen, and Zhan}]{cd_review_2020}
Shi W, Zhang M, Zhang R, et~al (2020) {Change Detection Based on Artificial
  Intelligence: State-of-the-Art and Challenges}. Remote Sensing 12(10)

\bibitem[{Silva~Junior et~al(2021)Silva~Junior, Pess{\^o}a, Carvalho, Reis,
  Anderson, and Arag{\~a}o}]{2020deforestation}
Silva~Junior CH, Pess{\^o}a AC, Carvalho NS, et~al (2021) {The Brazilian Amazon
  deforestation rate in 2020 is the greatest of the decade}. Nature Ecology \&
  Evolution 5(2):144--145

\bibitem[{Survey(2025)}]{landsat}
Survey USG (2025) {Landsat Project Documents}.
  \url{https://www.usgs.gov/landsat-missions/landsat-project-documents}

\bibitem[{Szegedy et~al(2015)Szegedy, Liu, Jia, Sermanet, Reed, Anguelov,
  Erhan, Vanhoucke, and Rabinovich}]{inception}
Szegedy C, Liu W, Jia Y, et~al (2015) {Going Deeper with Convolutions}. In:
  IEEE Conference on Computer Vision and Pattern Recognition, pp 1--9

\bibitem[{Szegedy et~al(2016)Szegedy, Vanhoucke, Ioffe, Shlens, and
  Wojna}]{rethinking_inception}
Szegedy C, Vanhoucke V, Ioffe S, et~al (2016) {Rethinking the Inception
  Architecture for Computer Vision}. In: IEEE Conference on Computer Vision and
  Pattern Recognition

\bibitem[{Taghanaki et~al(2019)Taghanaki, Zheng, {Kevin Zhou}, Georgescu,
  Sharma, Xu, Comaniciu, and Hamarneh}]{combo_loss}
Taghanaki SA, Zheng Y, {Kevin Zhou} S, et~al (2019) {Combo Loss: Handling Input
  and Output Imbalance in Multi-Organ Segmentation}. Computerized Medical
  Imaging and Graphics 75:24--33

\bibitem[{USGS(2025)}]{usgs}
USGS (2025) {United States Geological Survey}. \url{https://www.usgs.gov/}

\bibitem[{Young et~al(2017)Young, Anderson, Chignell, Vorster, Lawrence, and
  Evangelista}]{landsat_preprocessing}
Young NE, Anderson RS, Chignell SM, et~al (2017) {A Survival Guide to Landsat
  Preprocessing}. Ecology 98(4):920--932

\bibitem[{Zhang et~al(2025)Zhang, Zhang, Li, Shi, Li, and
  Zou}]{zhang2025biophysical}
Zhang Y, Zhang G, Li W, et~al (2025) {The Biophysical Effect of Loss of
  Different Forests on Land Surface Temperature in Idealized Deforestation
  Experiment}. Theoretical and Applied Climatology 156(4):1--19

\bibitem[{Zhou et~al(2018)Zhou, Rahman~Siddiquee, Tajbakhsh, and
  Liang}]{unet++}
Zhou Z, Rahman~Siddiquee MM, Tajbakhsh N, et~al (2018) {UNet\texttt{++}: A
  Nested U-Net Architecture for Medical Image Segmentation}. In: Deep Learning
  in Medical Image Analysis and Multimodal Learning for Clinical Decision
  Support. Springer International Publishing, Cham, pp 3--11

\end{thebibliography}

\end{document}